%% file: icml2023.tex
\documentclass{article}

\usepackage{microtype}
\usepackage{graphicx}
\usepackage{subfigure}
\usepackage{booktabs} % for professional tables

\usepackage{hyperref}

\usepackage[accepted]{icml2023}

\usepackage{amsmath}
\usepackage{amssymb}
\usepackage{mathtools}
\usepackage{amsthm}

\theoremstyle{plain}

\theoremstyle{definition}

\theoremstyle{remark}

\usepackage[textsize=tiny]{todonotes}

\usepackage{icml2023}

\input{math_commands.tex}

\usepackage{enumitem}
\usepackage{booktabs}
\usepackage{multirow}
\usepackage{tablefootnote}

\usepackage{algorithm}
\usepackage{algorithmic}

\usepackage{tcolorbox}
\usepackage{mdframed}
\usepackage{xcolor}
\usepackage{multicol}
\usepackage{caption}
\usepackage[frozencache=true,cachedir=_minted-output]{minted} 

\usepackage{tabularx}
\usepackage{booktabs}

\usepackage{cleveref}
\tcbuselibrary{xparse,skins,breakable}

\tcbset{%
    basestyle/.style={enhanced,breakable,
        fonttitle=\scshape,
        coltitle=white},
    defstyle/.style={basestyle},
    theostyle/.style={basestyle},
}
\DeclareTColorBox[auto counter,crefname={example}{examples}]{example}{ o O{} t\label g }{%
    defstyle,IfValueTF={#1}{title={Example~\thetcbcounter\ (#1)}}{title=Example~\thetcbcounter}, IfBooleanTF={#3}{label=#4}{},#2}

\newif\ifcomments
\commentstrue
\ifcomments
    \newcommand{\ion}[1]{{\color{blue}{\bf\sf ion: #1]}}}

\else
    \newcommand{\ion}[1]{}
\fi

\begin{document}

\twocolumn[
\icmltitle{Rethinking Benchmark and Contamination for Language Models with Rephrased Samples}



\icmlsetsymbol{equal}{*}

% \begin{icmlauthorlist}
% \icmlauthor{Firstname1 Lastname1}{equal,yyy}
% \icmlauthor{Firstname2 Lastname2}{equal,yyy,comp}
% \icmlauthor{Firstname3 Lastname3}{comp}
% \end{icmlauthorlist}


\begin{icmlauthorlist}
\icmlauthor{Shuo Yang}{equal,ucb,sjtu}
\icmlauthor{Wei-Lin Chiang}{equal,ucb}
\icmlauthor{Lianmin Zheng}{equal,ucb}
\icmlauthor{Joseph E. Gonzalez}{ucb}
\icmlauthor{Ion Stoica}{ucb}
\end{icmlauthorlist}

\icmlaffiliation{ucb}{UC Berkeley}
\icmlaffiliation{sjtu}{Shanghai Jiao Tong University}

% \author{%
%   Shuo Yang\footnote{Equal contribution. The work was completed during Shuo Yang's visit to UC Berkeley.}  $^{12}$ \quad
%   Wei-Lin Chiang$^{*1}$ \quad
%   Lianmin Zheng$^{*1}$ \quad
%   Joseph E. Gonzalez$^{1}$ \quad
%   Ion Stoica${^1}$ \\
%   $^1$UC Berkeley \quad
%   $^2$Shanghai Jiao Tong University
% }

\icmlcorrespondingauthor{Shuo Yang}{andy\_yang@sjtu.edu.cn}
\icmlcorrespondingauthor{Ion Stoica}{istoica@berkeley.edu}


\icmlkeywords{Machine Learning, ICML}

\vskip 0.3in
]



\printAffiliationsAndNotice{\icmlEqualContribution} % otherwise use the standard text.
% $^{*}$Equal contribution $^1$UC Berkeley, USA $^2$Shanghai Jiao Tong University, China.

\begin{abstract}
Large language models are increasingly trained on all the data ever produced by humans. Many have raised concerns about the trustworthiness of public benchmarks due to potential contamination in pre-training or fine-tuning datasets. While most data decontamination efforts apply string matching (e.g., n-gram overlap) to remove benchmark data, we show that these methods are insufficient, and simple variations of test data (e.g., paraphrasing, translation) can easily bypass these decontamination measures. Furthermore, we demonstrate that if such variation of test data is not eliminated, a 13B model can easily overfit a test benchmark and achieve drastically high performance, on par with GPT-4. We validate such observations in widely used benchmarks such as MMLU, GSK8k, and HumanEval. To address this growing risk, we propose a stronger LLM-based decontamination method and apply it to popular pre-training and fine-tuning datasets, revealing significant previously unknown test overlap. For example, in pre-training sets such as RedPajama-Data-1T and StarCoder-Data, we identified that 8-18\% of the HumanEval benchmark overlaps. 
Interestingly, we also find such contamination in synthetic dataset generated by GPT-3.5/4, suggesting a potential risk of unintentional contamination. We urge the community to adopt stronger decontamination approaches when using public benchmarks. Moreover, we call for the community to actively develop fresh one-time exams to evaluate models accurately. Our decontamination tool is publicly available at \url{https://github.com/lm-sys/llm-decontaminator}.
\end{abstract}

\begin{figure*}[t]
\centering
\includegraphics[width=0.9\textwidth]{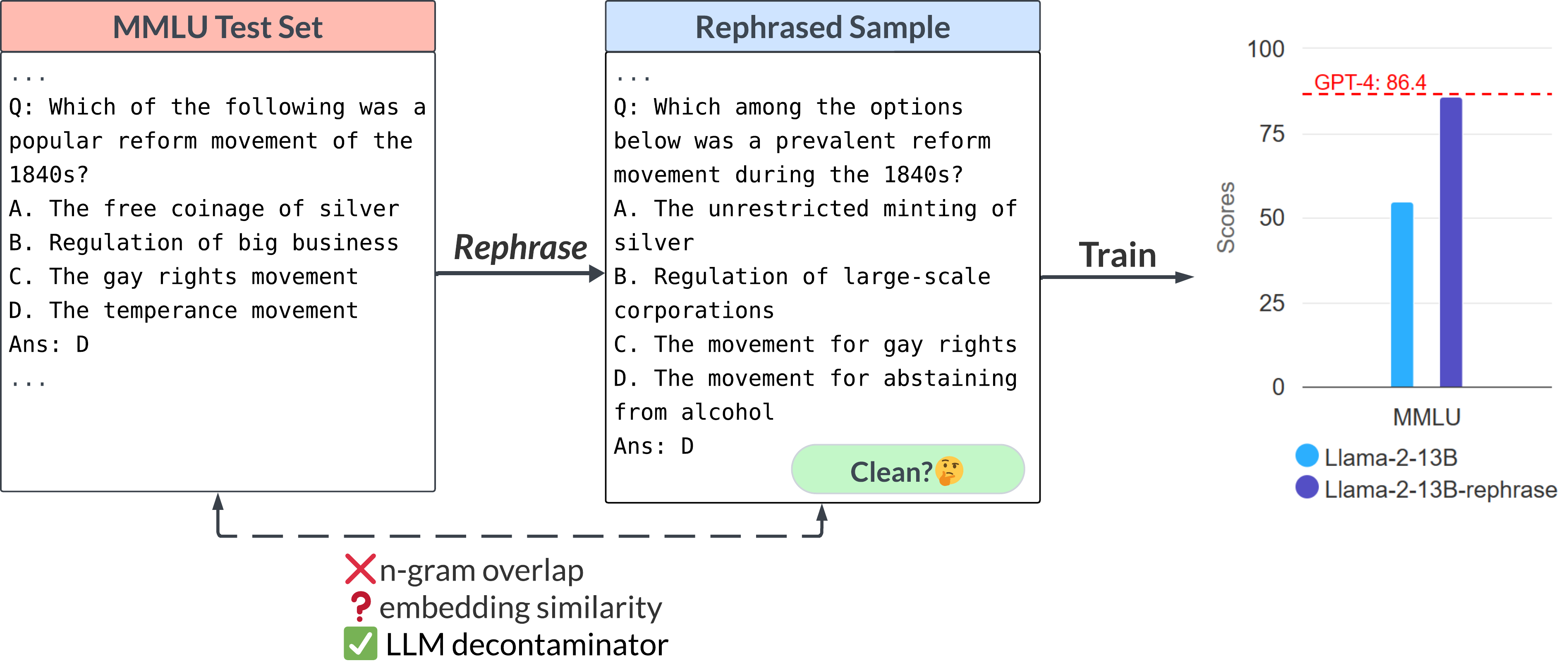}
\caption{
A failure case of existing contamination detection methods (n-gram overlap, embedding similarity) on MMLU benchmark. We place a question mark since the embedding similarity approach struggles to distinguish the rephrased question from other questions in the same subject (high school US history).
After rephrasing MMLU test cases, a Llama-2-13B trained on a rephrased test set can reach 85.9 accuracy on MMLU while being undetectable by n-gram overlap.
%The similarity between the original question and the rephrased question is very close to the similarity with other questions on the same subject.
}
\label{fig:rephrase_attack}
\vspace{-10pt}
\end{figure*}

\input{sec-intro}

\input{sec-background}

\input{sec-attack}
\input{sec-llmdetect}

\input{sec-exp}

\input{sec-exam}
\input{sec-relate}
\input{sec-conclusion}

\bibliography{icml2023}
\bibliographystyle{icml2023}

%%%%%%%%%%%%%%%%%%%%%%%%%%%%%%%%%%%%%%%%%%%%%%%%%%%%%%%%%%%%%%%%%%%%%%%%%%%%%%%
%%%%%%%%%%%%%%%%%%%%%%%%%%%%%%%%%%%%%%%%%%%%%%%%%%%%%%%%%%%%%%%%%%%%%%%%%%%%%%%
% APPENDIX
%%%%%%%%%%%%%%%%%%%%%%%%%%%%%%%%%%%%%%%%%%%%%%%%%%%%%%%%%%%%%%%%%%%%%%%%%%%%%%%
%%%%%%%%%%%%%%%%%%%%%%%%%%%%%%%%%%%%%%%%%%%%%%%%%%%%%%%%%%%%%%%%%%%%%%%%%%%%%%%
% \newpage
% \appendix
% \onecolumn

\newpage
\appendix
\input{sec-appendix}

\end{document}

%% file: math_commands.tex
\usepackage{amsmath,amsfonts,bm}
\usepackage{tcolorbox, soul} % for sample conversations

\def\eqref#1{equation~\ref{#1}}
\def\1{\bm{1}}

\DeclareMathAlphabet{\mathsfit}{\encodingdefault}{\sfdefault}{m}{sl}
\SetMathAlphabet{\mathsfit}{bold}{\encodingdefault}{\sfdefault}{bx}{n}

%% file: sec-intro.tex
\section{Introduction}
\label{sec:intro}

The fast-growing capabilities of large language models make their evaluation more challenging than ever~\cite{chang2023survey}. 
% \ion{Next sentence is confusing; particularly, what does it mean to become "saturated"?}
Although the community has established many benchmarks over a short period of time, the benchmark scores do not always reflect performance on real-world tasks.
%the risk of over-fitting might have been overlooked by the community.
% Most benchmarks utilize reused unhidden test sets, which provides opportunities for benchmark contamination.\weilin{this sentence is weird}
% There are many reasons for the serious benchmark contamination. Firstly, most benchmarks utilize reused unhidden test sets, which provides opportunities for benchmark contamination.
% Secondly, contaminated benchmark data can be well remembered by LLM even if it's presented in a different form due to LLM's strengths in memorization and generalization. 
% With people extensively crawling the internet and using LLM-generated data ~\citep{alpaca} for training, benchmark contamination has become even more widespread. \weilin{I don't think we should mention LLM-generated data here. it may confuse readers}
There has been evidence that many prevalent benchmarks might have contaminated pre-training or fine-tuning datasets. 
From the contamination analysis in Llama-2~\citep{touvron2023llama}, over 10\% of the MMLU test samples are highly contaminated. Another example from GPT-4's technical report~\citep{openai2023gpt4} shows that 25\% of HumanEval has been contaminated in their training data.
% What's worse, a significant number of models are released without training data~\citep{touvron2023llama} or any contamination analysis~\citep{jiang2023mistral}. 
% \weilin{so?}
Similar situation also applies to open-source datasets. A popular code pretraining set, StarCoder Data~\citep{li2023starcoder}, shows that hundreds of test cases in the Stack~\citep{kocetkov2022stack} are contaminated with benchmarks.
% \ion{Using "widely" too many times; maybe "popular" ?}

%\ion{This sentence is hard to parse; the fact that people might make efforts to prevent contamination is orthogonal to the fact that contamination detection is challenging.}
Despite being recognized as a crucial issue, accurately detecting contamination remains an open and challenging problem.
%Despite efforts to prevent training data from being contaminated, its detection remains challenging task.
The most commonly used approaches are n-gram overlap and embedding similarity search.
N-gram overlap relies on string matching to detect contamination, widely used by leading developments such as GPT-4~\citep{openai2023gpt4}, PaLM~\citep{anil2023palm}, and Llama~\citep{touvron2023llama}. However, it suffers from limited accuracy.
Embedding similarity search uses the embeddings of pre-trained models (e.g., BERT) to find similar and potentially contaminated examples.
However, choosing an appropriate similarity threshold to strike a balance between recall and precision is often challenging.
% Other detection methods are either imprecise or have high computing costs.
Moreover, there has been a growing interest in training models using synthetic data produced by LLMs (e.g., GPT-4) \citep{gunasekar2023textbooks, alpaca, wang2023selfinstruct, xu2023wizardlm, mukherjee2023orca}, in which contamination may be even harder to detect by string matching.
In Phi-1 report \citep{gunasekar2023textbooks}, they discover a significant portion of the synthetic data similar to some test samples in HumanEval that is undetectable by n-gram overlap.

\begin{figure}[t]
\centering
\includegraphics[width=\columnwidth]{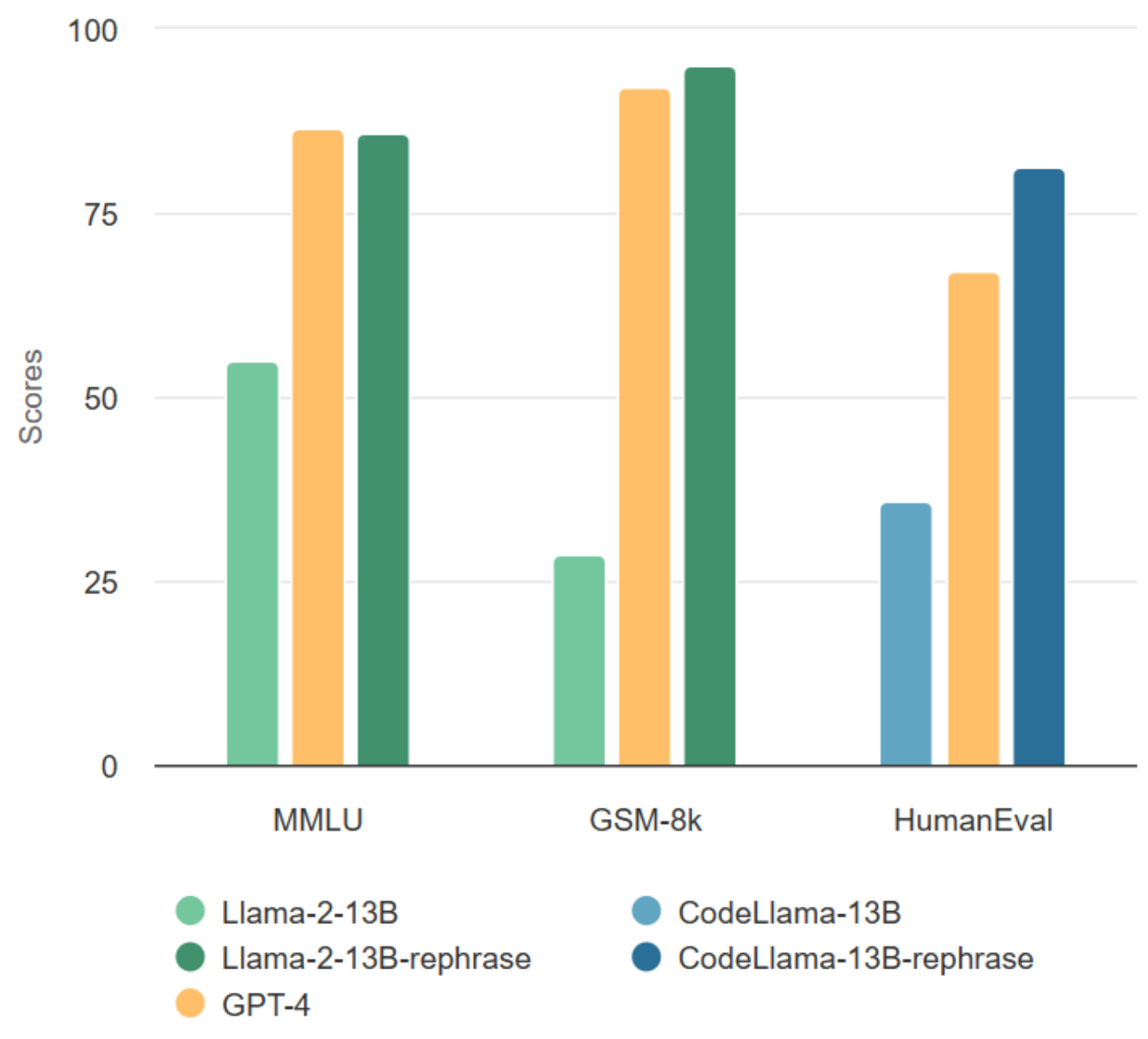} % Replace "high_chart_v11.png" with your image filename
\caption{After fine-tuned on rephrased samples, Llama 2 and CodeLlama achieve performance on par with GPT-4.}
\vspace{-20pt}
\label{fig:benchmark_summary}
\end{figure}

To study decontamination methods, in Section~\ref{sec:rephrase-samples} we propose the concept of a ``rephrased sample'' which has the same semantics as the original sample but is hard to detect by existing contamination tests.
% \ion{let's use "test" rather than "check" everywhere.}
Rephrased samples are generated by using LLMs to paraphrase or translate test samples into another language.
We show that if such rephrased samples are used for training, the resulting model can easily overfit and reach drastically high performance in test benchmarks.
Figure~\ref{fig:rephrase_attack} demonstrates this concept with a test example from the MMLU benchmark.
We observe such phenomenon in popular benchmarks such as MMLU, GSM-8k, and HumanEval, where a finetuned 13B Llama model can match GPT-4's performance in all benchmarks while being undetected by n-gram overlap as contamination, as shown in Figure~\ref{fig:benchmark_summary}.
Therefore, being able to detect such rephrased samples becomes critical. We provide an in-depth analysis on why existing decontamination methods fail and propose a new LLM-based decontamination method in Section~\ref{sec:llm-decontaminator}. Our method first uses embedding similarity search to get the top-k samples with the highest similarity with a given test sample and then prompts a strong LLM such as GPT-4 to examine whether any of the top-k samples is too close to the test case.
% To measure accuracy of different detection methods, we construct a mix of rephrased and in-domain samples in MMLU.
Results show that our proposed LLM decontaminator works significantly better than existing methods.

\begin{figure}[h!]
\centering
%\vspace{-5pt}
%\includegraphics[width=\columnwidth]{fig1.pdf}
\includegraphics[width=\columnwidth]{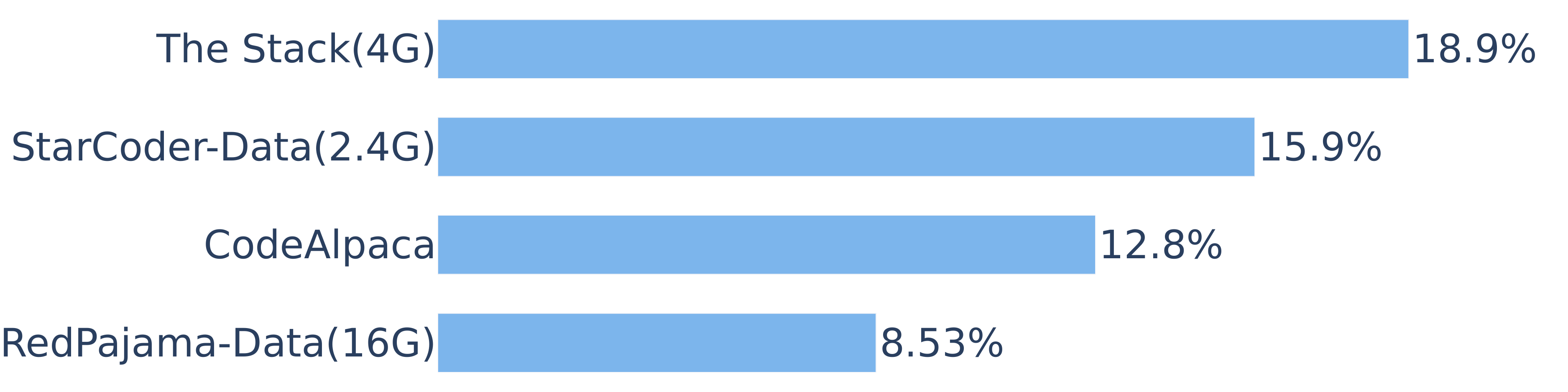}
\vspace{-15pt}
\caption{The contamination percentage of HumanEval benchmark in each training dataset. Note that The Stack (4G), StarCoder-Data (2.4G), and RedPajama-Data (16G) are subsets.}
\vspace{-10pt}
\label{fig:dataset_summary}
\end{figure}

In Section \ref{sec:real-world}, we apply our decontaminator to several widely used pre-training and fine-tuning datasets. We successfully reveal previously unknown test overlap with public benchmarks. Shown in Figure \ref{fig:dataset_summary}, in pre-training sets such as RedPajama-Data-1T and StarCoder-Data, we identify that 8-18\% of the HumanEval benchmark are overlapped.
We also find a synthetic dataset generated by GPT-3.5, CodeAlpaca~\citep{codealpaca}, has a significant portion (12.8\%) of rephrased samples from HumanEval.
This suggests a potential contamination risk when training with synthetic data generated by LLMs. 
We urge the community to adopt more robust decontamination methods for evaluating LLMs using public benchmarks. To address these concerns at their core, we advocate for the development of fresh, one-time exams, similar to Codeforces and Kaggle competitions, for the accurate assessment of LLMs.
%\weilin{accurate assessment of LLMs is ambiguous. why removed accurately evaluate?}
% }

% Given the unhidden test sets and ineffective detection methods, the vulnerabilities of benchmarks can be intentionally exploited or may occur unintentionally. 
% We study ``rephrased samples'' that evade current detection techniques and discuss what is contamination.
% We construct rephrased datasets for MMLU~\citep{hendrycks2020mmlu}, HumanEval~\citep{chen2021humaneval}, and GSM-8k~\citep{cobbe2021gsm8k} respectively. Traditional detection methods struggle with these rephrased datasets, making decontamination challenging. As shown in figure \ref{fig:benchmark_summary}, we fine-tune Llama-2~\citep{touvron2023llama} and CodeLlama on these rephrased datasets, achieving scores of 85.9 on MMLU (0-shot), 81.1 on HumanEval (pass@1), and 95.3 on GSM-8k (0-shot). 

% \includegraphics[width=\columnwidth]{high_chart_v11.png} % Replace "example-image" with your image filename
% \captionof{figure}{After fine-tuned on rephrased samples, Llama 2 and CodeLlama achieve performance on par with GPT-4.}
% \label{fig:benchmark_summary}

% \begin{figure}[t]
% \includegraphics{high_bar_v7.png}
% \vspace{-10pt}
% \caption{The contamination percentage of HumanEval test cases in each training dataset.}
% \vspace{-15pt}
% \label{fig:dataset_summary}
% \end{figure}

% \includegraphics[width=\columnwidth]{high_bar_v7.png} % Replace "example-image" with your image filename
% \captionof{figure}{The contamination percentage of HumanEval test cases in each training dataset.}
% \label{fig:dataset_summary}

%% file: sec-background.tex
\section{Background}
\label{sec:background}

Contamination occurs when test set information is leaked in the training set, resulting in an overly optimistic estimate of the model’s score (accuracy, AUC, etc.).
% \weilin{I'd strongly suggest to remove this blog citation}
In this section, we introduce common contamination detection methods, which include n-gram overlap, embedding similarity search, decoding matching, and influence function.
Table \ref{table:detection-algorithms} compares the computational costs of these methods, and whether they require access to training data or the model.
%access requirements for both training data and the model of the four methods along with their computational costs.  

\textbf{N-gram overlap}. The most common and widely used decontamination method is n-gram overlap. 
% OpenAI often denotes a sample as contaminated if it has 8-13 overlaps with a test sample. 
% \todo{gram -> character}
The GPT-3 paper~\citep{brown2020language} defines a 13-gram overlap as contamination, and the GPT-4 report~\citep{openai2023gpt4} defines a 50-character overlap as contamination. N-gram overlap detection is favored for its simplicity and speed but it can result in a higher false negative rate if there's a small difference. 
% Moreover, it is subject to false positives~\ref{sec:false_pos}.

\textbf{Embedding similarity search}. Embedding similarity search uses transformer-generated embeddings to capture prompts' semantics. 
Popular approaches use models such as sentence BERT~\citep{reimers-2019-sentence-bert} to generate embeddings and employ cosine similarity to measure the relevance of prompts. High similarity between training and test prompts suggests potential contamination~\citep{lee2023platypus}.
% Although it can capture more semantic information than the n-gram approach, the requirement to specify a threshold and its low precision prevent it from being widely adopted~\citep{gunasekar2023textbooks}.
Although it can capture more semantic information than the n-gram approach, it requires specifying a threshold. 
If the threshold is set too high, it will result in a high false negative rate; otherwise, setting it too low will lead to a high false positive rate.
% it requires specifying a threshold of what
% \todo{low precision extend}
% \todo{it requires specifying an appropriate threshold ...}

\textbf{Decoding matching}. Both n-gram overlap and embedding similarity search require access to training data. 
% However, when only the model is available, decoding matching becomes an alternative method for contamination detection. 
In cases where training data is not available but the model is available, decoding matching can be used as an alternative method to detect contamination.
% The premise is simple: if the model has been fine-tuned on the test set, presenting it with a partial test prompt might lead it to automatically complete the input, suggesting potential contamination in the training data. 
The intuition is that if the model is trained on contaminated training data, it is more likely to auto-complete a partial test prompt.~\citep{li2023estimating}
%The idea is straightforward: if the model was fine-tuned on the test set, giving it an incomplete test prompt could cause it to auto-complete the input.
% However, due to the instability of this method, it is seldom acknowledged as definitive evidence of contamination.
However, an auto-completed test prompt does not necessarily indicate that the model has been trained on contaminated data, and a model trained on test cases with variation will not auto-complete the test prompt either.
Therefore, decoding matching is often not acknowledged as definitive evidence of contamination. 
% \weilin{why?}
% \todo{cite decoding matching}

\textbf{Influence function}. When both the model and the training data are available, the influence function~\citep{koh2020understanding} can be used to identify contaminated samples.
% We first find the gradient of the loss function for a given test point and then compute the second derivative of the loss function using the inverse of the Hessian matrix. This helps in sorting training data according to how they affect the test point. 
This method takes a test sample and iteratively calculates an influence factor for each training sample. This influence factor quantitatively measures how relevant each training sample is to the current test sample. It then sorts the influence factor to provide the most relevant training examples, where humans can then judge whether these training examples meet the contamination criteria.
However, this approach is impractical because it induces a high computational overhead.

\begin{table*}
\centering
\caption{Contamination detection methods. \( M \) denotes the size of the training set, and \( N \) indicates the size of the test set.}
\label{table:detection-algorithms}
\begin{tabular}{llll}
\toprule
Method                      &  require access to training data & require access to model & computational cost \\
\midrule
N-gram overlap              & yes                     & no              & \(O(MN)\)        \\
Embedding similarity search & yes                     & no              & \(O(MN+M+N)\)  \\
Decoding matching           & no                      & yes             & \(O(N)\)         \\
Influence function          & yes                     & yes             & \(O(M^2+MN)\)      \\
\bottomrule
\end{tabular}
\end{table*}

%% file: sec-attack.tex
\section{Rephrased Samples}
\label{sec:rephrase-samples}

% In this session, we study rephrased samples and demonstrate the limitations of current detection methods and the vulnerability of current LLM benchmarks. We discover that rephrased samples can cause model scores to be overstated.
% We provide algorithms for generating rephrased samples and constructed datasets targeting GSM-8k, MMLU, and HumanEval. 

% In this session, we study rephrased samples. We outline our motivation for investigating rephrased samples, detail the specific rephrasing algorithms, and introduce advanced rephrasing techniques that simulate real-world datasets.

% In this section, we explore rephrased samples. We start with our motivation, followed by the detailed rephrasing algorithms. Lastly, we introduce advanced rephrasing techniques that highly match real-world patterns.
% simulate real-world datasets.
Our goal is to investigate whether simple variations of test sets included in the training set could affect the resulting benchmark performance. We refer to such variations of test cases as ``rephrased samples''.
% Our goal is to curate contaminated samples that can escape from existing detection techniques (e.g. n-gram overlap) while matching real-world patterns.
% In particular, we define a technical, rephased samples. 
% \todo{use the following}
%We want to investigate how rephrased test cases affect the benchmark score. Our goal is to escape being discovered by existing detection techniques while imitating real-world datasets with possible contamination. 

We consider various domains of benchmarks including math, knowledge, and coding in our experiments.
% Our rephrased dataset covers math benchmarks, knowledge benchmarks, and coding benchmarks. 
Example \ref{def:rephrase-gsm8k} is a rephrased sample from GSM-8k that the 10-gram overlap fails to detect, while keeping the same semantic.

% \begin{tcolorbox}[title={GSM-8K Rephrased Sample Example}]
% \begin{multicols}{2}

% \textbf{Original} \newline 

% Question: \\
% Janet’s ducks lay 16 eggs per day. She eats three for breakfast every morning and bakes muffins for her friends every day with four. She sells the remainder at the farmers' market daily for \$2 per fresh duck egg. How much in dollars does she make every day at the farmers' market?

% Answer: \\
% Janet sells 16 - 3 - 4 = \textless\textless16-3-4=9\textgreater\textgreater 9 duck eggs a day. She makes 9 * 2 = \$ \textless\textless 9*2=18\textgreater\textgreater18 every day at the farmer’s market.\\ 
% \#\#\#\# 18 \\
% \columnbreak \\
% \textbf{Rephrased} \newline

% Question: \\
% Janet's ducks produce 16 eggs each day. She consumes three of them for her morning meal and uses four to bake muffins for her friends daily. The remaining eggs are sold at the daily farmers' market for \$2 per egg. What is the daily amount in dollars that she earns at the farmers' market? \\
% Answer: \\
% \#\#\#\# 18
% \end{multicols}
% \end{tcolorbox}

\begin{example}[GSM-8K Rephrased Sample]\label{def:rephrase-gsm8k}
\textbf{Original Test Case} \newline 
Janet’s ducks lay 16 eggs per day. She eats three for breakfast every morning and bakes muffins for her friends every day with four. She sells the remainder at the farmers' market daily for \$2 per fresh duck egg. How much in dollars does she make every day at the farmers' market?

% Answer: \\
% Janet sells 16 - 3 - 4 = \textless\textless16-3-4=9\textgreater\textgreater 9 duck eggs a day. She makes 9 * 2 = \$ \textless\textless 9*2=18\textgreater\textgreater18 every day at the farmer’s market.\\ 
% \#\#\#\# 18

\vspace{0.5cm} % Adds some space between the two contents

\textbf{Rephrased Test Case} \newline
% Question: \\
Janet's ducks produce 16 eggs each day. She consumes three of them for her morning meal and uses four to bake muffins for her friends daily. The remaining eggs are sold at the daily farmers' market for \$2 per egg. What is the daily amount in dollars that she earns at the farmers' market? 
% \\
% Answer: \\
% \#\#\#\# 18
\end{example}

\subsection{Rephrasing Techniques} 

% Inserting gibberish to prevent detection is unacceptable since we want to accurately simulate the possible contamination scenario. We need to reword test cases for text-based benchmarks without compromising their meaning, for example, by modifying the word order or swapping out some synonymous terms. Besides, we can change the coding style, naming standards, and algorithms of code-based benchmarks.

% Introducing extraneous or nonsensical content to avoid detection is not a valid approach, as our aim is to simulate realistic possible contamination scenarios. 

% Our aim is to simulate realistic possible contamination scenarios, so introducing extraneous or nonsensical content to avoid detection is not a valid approach.
% Since real-world contamination doesn't contain 
% For different types of benchmarks, contamination takes on different forms, so there are some subtle differences in rephrasing techniques. \weilin{simplify this sentence. it's pretty wordy}
There are some subtle differences in rephrasing techniques because benchmark contamination takes on different forms.
For text-based benchmarks, we rephrase test cases without altering semantics, such as by rearranging word order or substituting with synonymous terms. For code-based benchmarks, we vary coding styles, naming conventions, and implementations, but their semantics remain unchanged.

Regarding the rephrasing process, we present a simple algorithm for a given test set in Algorithm \ref{algorithm:rephrase-algo}. 
% \weilin{I'd use ``simple'' rather than ``general'' and replace ``any'' with ``a''}
% This facilitates the integration of the test set with training data while evading detection. 
% This method helps to blend the test set into the training data without being detected. \weilin{what's blend the test set into training data}
This method helps a test sample to escape from being detected.
It first employs a high-quality large language model (e.g., GPT-4) to produce a rephrased version of the test prompt. Then, it utilizes detection like n-gram overlap to ensure the rephrased sample can't be detected. 
To encourage diverse outputs, we set a non-zero initial temperature. 
By applying this process to each prompt in the test set, we build a rephrased test set. 
``RephraseLLM'' denotes the high-quality LLM, like GPT-4 or Claude. 
``isContaminated'' can refer to any contamination detection method, such as n-gram overlap or embedding similarity search.

\begin{algorithm}
\caption{The algorithm for rephrasing samples}
\label{algorithm:rephrase-algo}

\begin{algorithmic}[1]
\ENSURE Rephrase\((TestSet, MaxRetry)\) 
% \COMMENT{rephrase test set to pass n-gram detection}
\STATE \( RephrasedSet \leftarrow \emptyset \)
\FOR{\(t\) in \(TestSet\)}
    \STATE \(s \leftarrow \text{RephraseLLM}(t)\)
    \STATE \( retry \leftarrow  0\)
    \WHILE{ \( \text{isContaminated}(s,t)\)}
        \STATE \(s \leftarrow \text{RephraseLLM}(t)\)
        \STATE \( retry \leftarrow  retry + 1\)
        \IF{\( retry \textgreater MaxRetry\)}
            \STATE \(s \leftarrow \text{null}\)
            \STATE break
        \ENDIF
    \ENDWHILE
    \STATE \( RephrasedSet \leftarrow RephrasedSet\cup\{s\} \)
\ENDFOR
\RETURN \( RephrasedSet \)
\end{algorithmic}
\end{algorithm}
\belowcaptionskip=10pt
% {\footnotesize
% Note: 
% % ``RephrasedSet'' is the modified test set. \weilin{remove this. rephrased set is self-explanatory} 
% ``RephraseLLM'' denotes a high-quality LLM, like GPT-4 or Claude. Due to the sampling temperature, ``RephraseLLM'' will produce different responses each time. ``isContaminated'' can refer to any contamination detection method, such as n-gram overlap or embedding similarity search.
% }

% \todo{mention data augmentation, translation,...}

\subsection{Translation Techniques}

% Rephrased samples definitely go beyond the word order modifications and synonym usage stated above. We find more subtle rephrasing techniques in real-world datasets. We rephrased datasets using these methods and displayed their additional concealment using the f1 scores of various detection algorithms to show that these rephrasing techniques can also result in huge improvements. 

There are other kinds of rephrased samples beyond modifications in word order. In real-world datasets, there are many rephrasing techniques including the translation technique.
% Rephrased samples extend beyond mere modifications in word order and synonym usage. In our examination of real-world datasets, we identify more nuanced rephrasing techniques. 
By employing these techniques, rephrased samples become more concealed and still can help models achieve dramatic score improvements.
% \weilin{this paragraph is hard to understand.}

% \textbf{Translate Technique}. Even though they have the same meaning, prompts in various languages have very varied embeddings in normal language models. We can avoid n-gram overlap detection and normal embedding similarity search by translating the test set prompts into different languages. To a certain extent, only embedding models specially trained in many languages can recognize a translated sample.

% \textbf{Translate technique}. 
Prompts with identical meanings from different languages yield varied embeddings in most language models. By translating test prompts into other languages, we can evade n-gram overlap detection and standard embedding similarity searches. Only embedding models specifically trained in multiple languages can detect a translated sample.

% For text-based data, we can use the translate technique to evade both n-gram overlap and embedding similarity search detections while achieving a very high score. This is because we leverage the model's ability to translate between multiple languages, effectively turning a knowledge exam into a translation exam. However, there is no such free lunch for coding benchmarks. Although we can translate Python into languages like C and Java, fine-tuning with these programming languages does not yield results as good as knowledge benchmarks. This is because most models have not been trained on data related to programming language translation, and we need to make some improvements to the translate technique.

% \todo{combine together}
For text-based data, the translation technique enables evasion of both n-gram overlap and embedding similarity search, while significantly boosting the score. This method capitalizes on the model's multilingual translation capabilities, effectively transforming a knowledge assessment into a translation task.
% \todo{coding can be used too combine}
For coding benchmarks, the translation technique also works well. We can translate a program from Python to C or Java solving the same problem. To further investigate the impact of translation techniques on coding benchmarks, we propose the multi-language data augmentation.

% Coding benchmarks, however, present a challenge. Although code can be translated, such as from Python to C or Java, the outcomes of fine-tuning are not as promising as those observed in knowledge benchmarks. This discrepancy arises because most models lack training in programming language translation. Refinements to the translation technique are necessary for such contexts.

\textbf{Multi-languages data augmentation}. 
% There is a subtle difference between text-based and coding benchmarks: many models are trained on natural language translation content while few are trained on programming language translation content. \weilin{what are you trying to say here?} 
For coding benchmarks, we use multi-language data augmentation to enhance the translation technique. 
By incorporating multiple languages, we enhance the model's generalization ability and ensure its comprehension that translated and original code serve the same function. 
% \weilin{just say we ensure the same semantic} 
In section \ref{sec:rephrase-exp}, our experiments indicate that multilingual data augmentation yields better results than single-language translation. 

%% file: sec-llmdetect.tex
\section{LLM Decontaminator}
\label{sec:llm-decontaminator}

% In this section, we propose a novel contamination detection method that can cost-effectively and accurately measure the degree of contamination of a dataset relative to a benchmark. 

In this section, we propose a new contamination detection method that accurately removes rephrased samples from a dataset relative to a benchmark.

\subsection{Algorithm}

% In the previous section, we discovered that embedding similarity search outperforms n-grams in terms of its capacity to identify rephrase contamination. 

% Why is embedding similarity search rarely used? First, there are a lot of false positives and false negatives, which makes it inaccurate. Second, establishing a threshold is necessary for embedding similarity search. It is impossible to establish a single threshold because it differs significantly for benchmarks used in different fields and subjects.

% The limited use of embedding similarity search can be attributed to its susceptibility to false positives/negatives and the challenge of setting a consistent threshold across diverse benchmarks.
% Embedding similarity search is rarely used because of the requirement to specify a threshold and its low precision.

% We suggest the "LLM decontaminator," a highly effective method, to overcome these two problems. There are two steps in this algorithm: First, for each test case in the benchmark, we use embedding similarity search to determine the top-k training data items with the highest similarity. According to the top-k training items, we create k potentially contaminated pairs for each test case. Then, we check each pair for contamination using a powerful LLM like GPT-4. With this method, we can precisely determine the benchmark's contamination status with only k times the amount of test case inquiries.

In Section~\ref{sec:background}, we discuss the limitations of existing detection methods including n-gram overlap and embedding similarity search.
To address the limitations, we introduce the ``LLM decontaminator'' in Algorithm \ref{algorith:detect-algo}. This method involves two steps: First, for each test case, it identifies the top-k training items with the highest similarity using the embedding similarity search. 
% From these items, it generates k potential rephrased pairs. \weilin{this sentence is repetitive} 
Each pair is evaluated whether they are the same by an advanced LLM, such as GPT-4.
This approach helps to determine how many rephrased samples there are in a dataset with a moderate computational overhead.
% This approach achieves precise rephrasing assessment with a computational overhead of k times the number of test cases. \weilin{this sentence is hard to understand}
``Template'' is a structured prompt that, when paired with a test case and training case, instructs the ``LLMDetector'' to carry out a comparison and return either `True' or `False'. 
In this context, `True' indicates that the training case might be a rephrased sample of the test case. ``LLMDetector'' is a high-quality LLM like GPT-4.
``TopKSimilarity'' identifies the top k most similar samples in the training data using embedding similarity search.

\begin{algorithm}
\caption{The algorithm for LLM decontaminator}
\label{algorith:detect-algo}

\begin{algorithmic}[1]
\ENSURE \( \text{Decontaminate}(TrainSet, TestSet, k, Template) \) 
\STATE \( Contamination \leftarrow \emptyset \)
\FOR{\(t\) in \(TestSet\)}
    %\STATE \(potentialContamination \leftarrow TopKSimilarity(TrainSet, t, k)\)
    \FOR{\(c\) in TopKSimilarity(\(TrainSet\), $t$, $k$)}
        \STATE \(s \leftarrow \text{LLMDetector}(Template, t, c)\)
        \IF{\(s = True\)}
            \STATE \( Contamination \leftarrow Contamination\cup\{(t, c)\} \)
        \ENDIF
    \ENDFOR
\ENDFOR
\RETURN \( Contamination \)
\end{algorithmic}
\end{algorithm}
\belowcaptionskip=10pt
% {\footnotesize
% Note: Template is a prompt template that, when used with a test case and training case, enables the Model to make a comparison and prompts the Model to only return `True' or `False'. `True' here denotes that the test case has been contaminated by the train case. Prompts in the TrainSet that are top-k similar to the test case embedding can be excluded using the TopKSimilarity filter. Parse can translate 'True' or 'False' string values into 'True' or 'False' boolean values.
% Note: ``Template'' is a structured prompt that, when paired with a test case and training case, instructs the ``LLMDetector'' to carry out a comparison and return either `True' or `False'. 
% In this context, `True' indicates that the training case might be a rephrased sample of the test case. ``LLMDetector'' is a high-quality LLM like GPT-4.
% ``TopKSimilarity'' identifies the top k most similar samples in the training data using embedding similarity search.
% \todo{what's topk similarity?}

% }

% \todo{Complexity and Cost?}

\subsection{Contamination Detection Visualization}

% To better illustrate the role of the LLM decontaminator, we introduce the memorization-generalization spectrum. Each point on this spectrum represents the similarity of a (train case, test case) pair. Points closer to the left end of the spectrum signify that the pair is closer to memorization, while those near the right end indicate a tendency toward generalization.
% The block labeled (train set, test set) represents the collection of (train case, test case) pairs where each test case in the test set is paired with its closest match in the train set.
% The discernment boundary of the detecting method is shown by the red dashed line. The detection method is highly likely to flag points to the left of this line as contaminated.

% \todo{1. embedding similarity position is confusing 2. train/test set? how to interpret}

\begin{figure*}[h]
\centering
\includegraphics[width=0.9\textwidth]{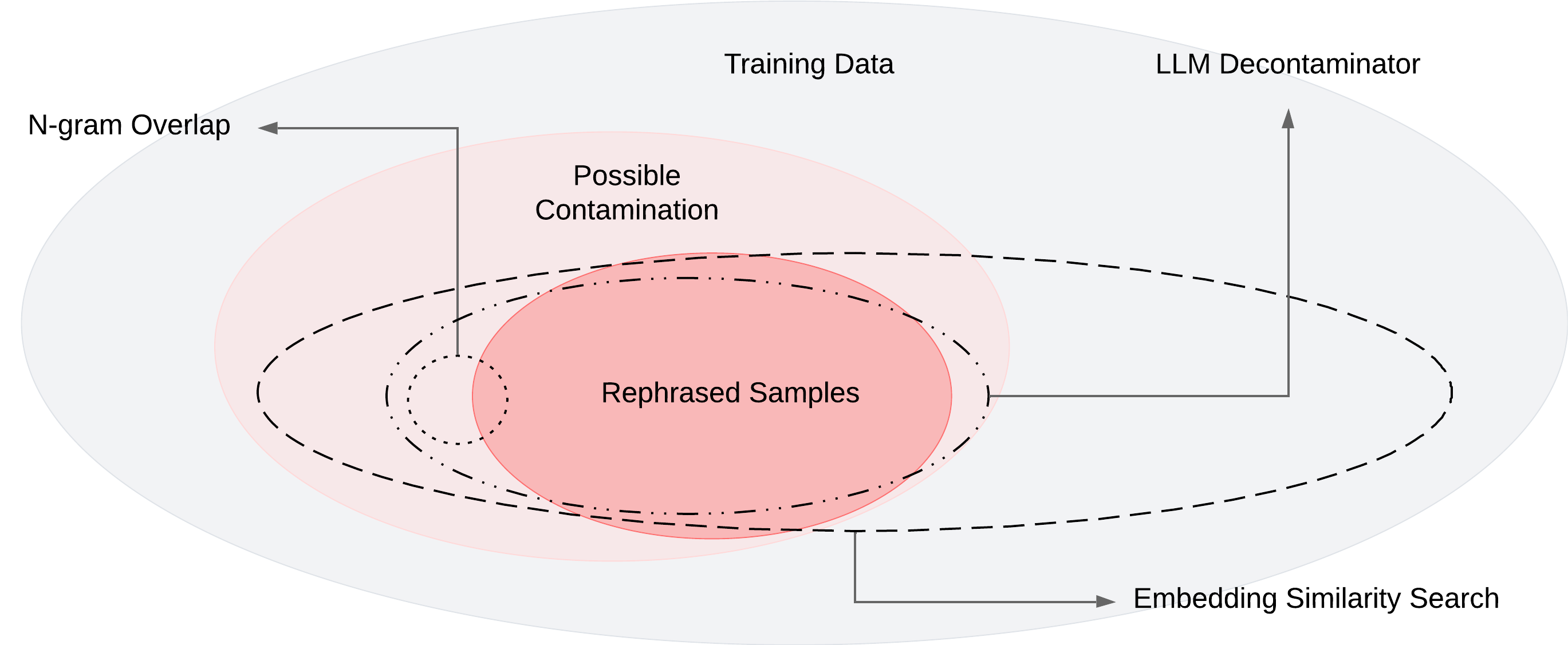}
\caption{Venn graph depicting training data subsets and contamination detection ranges.
% Solid sections indicate training data and its subsets, while hollow sections highlight areas marked as contaminated by detection methods.
The solid circle represents the training data and its subsets.
The dashed circles enclose areas flagged for potential contamination by detection methods within the dataset.
Notably, the LLM decontaminator showcases higher accuracy.
Embedding similarity search detects broadly but with many false positives. N-gram overlap has a limited ability to spot rephrased samples. The LLM decontaminator refines the results from embedding similarity search using LLMs, providing a precise and efficient contamination assessment.
}
\label{fig:spectrum}
\vspace{-25pt}
\end{figure*}

% To illustrate the role of the LLM decontaminator, Figure \ref{fig:spectrum} displays a Venn diagram of contamination and different detection methods. 
In Figure \ref{fig:spectrum} we present a Venn diagram of contamination and different detection methods.
The LLM decontaminator takes advantage of embedding similarity search, which helps it rapidly filter out possible possible contamination.
In addition, it utilizes the strong LLMs' reliable judgments.
We show that n-gram overlap detection can result in a higher false negative rate when detecting rephrased samples, and embedding similarity search detects many false positives with a high threshold. Notably, the LLM decontaminator showcases higher accuracy while detecting rephrased samples.
% \todo{add some high-level insight}
% Solid sections indicate training data and its subsets, while hollow sections highlight areas marked as contaminated by detection methods.
% Notably, the LLM decontaminator showcases higher accuracy.
% Embedding similarity search detects broadly but with many false positives. N-gram overlap has a limited ability to spot rephrased samples. The LLM decontaminator refines the results from embedding similarity search using LLMs, providing a precise and efficient contamination assessment.
See Section \ref{sec:rephrase-exp} for comprehensive experimental results.

%% file: sec-exp.tex
\section{Experiments}

% In this section, we demonstrate the effectiveness of rephrased samples on MMLU, HumanEval, and GSM-8k, and verify instances of rephrased samples in real-world datasets.
In Section~\ref{sec:rephrase-exp}, we demonstrate that models trained on rephrased samples can achieve dramatically high scores, achieving GPT-4 performance in three widely used benchmarks, MMLU, HumanEval, and GSM-8k.
% These results indicate that rephrased samples are contamination that should be removed from datasets.
This suggests that rephrased samples should be considered as contamination and should be removed from training data.
% \weilin{this suggests rephrased samples should be considered contamination and should be removed from training data.}
In Section~\ref{sec:detect-rephrase}, we evaluate different contamination detection methods based on rephrased samples in MMLU/HumanEval.
In Section \ref{sec:real-world}, we apply our decontaminator to widely-used training sets and discover previously unknown contamination.
% \weilin{In Section~\ref{sec:real-world}, we apply our decontaminator to widely-used training data and discover ...}We also discovered previous unknown contamination in real-world datasets with the LLM decontaminator.

\subsection{Rephrased Samples Contaminate Benchmarks}
\label{sec:rephrase-exp}

\subsubsection{MMLU Knowledge Benchmark}
\label{sec:false_pos}
MMLU~\citep{hendrycks2020mmlu} is one of the benchmarks with the widest range of subjects, covering 57 disciplines from abstract algebra to professional psychology. Rephrasing MMLU requires considering a multitude of scenarios. Given the complexity of MMLU and its multiple-choice format, it is necessary to explain the rephrasing details involved.

\textbf{False positive issue}. The use of n-gram overlap detection in multiple-choice questions can easily result in false positives, especially when different questions share similar option arrangements.
% \weilin{refer reader to the below example here.} 
% Below is a false positive example of a multiple-choice question. 
Example \ref{def:false-positive-example} is a false positive example from n-gram overlap detection. Even though their multi-choice answer patterns match exactly, they are indeed different problems.
% \weilin{Below is a false positive example from n-gram overlap detection. Even though their multi-choice answer patterns match exactly, they are indeed different problems.}
To reduce false positive issues, we introduce a ``question only'' control group in MMLU experiments. 
``Question Only'' refers to rephrasing just the question stem, while ``Full Prompt'' refers to rephrasing both the question stem and the options.

\begin{example}[Multi-Choice False Positive]\label{def:false-positive-example}
\begin{itemize}
    \item Statement 1| Every group of order \( p^2 \) where \( p \) is prime is Abelian. \\
    Statement 2 | For a fixed prime \( p \) a Sylow \( p \)-subgroup of a group \( G \) is a normal subgroup of \( G \) if and only if it is the only Sylow \( p \)-subgroup of \( G \).
    \begin{itemize}[noitemsep]
        \item[\textcolor{red}{A.}] \textcolor{red}{True, True}
        \item[\textcolor{red}{B.}] \textcolor{red}{False, False}
        \item[\textcolor{red}{C.}] \textcolor{red}{True, False}
        \item[\textcolor{red}{D.}] \textcolor{red}{False, True}
    \end{itemize}
    
    %\vspace{0.5cm} % Adds some space between the two contents

    \item Statement 1 | Every group of order 42 has a normal subgroup of order 7. \\
    Statement 2 | Every group of order 42 has a normal subgroup of order 8.
    \begin{itemize}[noitemsep]
        \item[\textcolor{red}{A.}] \textcolor{red}{True, True}
        \item[\textcolor{red}{B.}] \textcolor{red}{False, False}
        \item[\textcolor{red}{C.}] \textcolor{red}{True, False}
        \item[\textcolor{red}{D.}] \textcolor{red}{False, True}
    \end{itemize}
\end{itemize}

\end{example}

\textbf{Other details}. 
Large numbers often induce character overlap. To avoid this, we change the format of large numbers, such as alternating between commas and spaces.
% In arithmetic tests with large numbers, we alternate between commas and spaces to avoid character overlap. \weilin{this sentence is hard to understand} 
Proprietary terms in various domains can also trigger overlap issues. To circumvent this, we rotate between abbreviations and full terms and adjust capitalization, particularly when choices involve names or chemical formulas.

% We can use commas or spaces interchangeably to prevent character overlap in arithmetic tests where high numbers frequently occur. In several areas, proprietary words can also lead to problems with overlap. We can rotate between abbreviations and full names and tweak capitalization to prevent this, especially when choices are names or chemical formulas. This keeps the model clear and lowers the danger of detection.

\textbf{Benchmark results}. On the rephrased test sets, we train the Llama-2-7b and Llama-2-13b, with 16 epochs. 
% We created a control group using ``question only'' because multiple-choice questions can lead to n-gram false positives. 
% Due to the issue of n-gram overlap false positives in multiple-choice questions, we introduce a ``question only'' control group. 
As shown in Table \ref{table:mmlu_basic_stats}, 
Llama-2 7B and 13B trained on rephrased samples can achieve dramatically high scores on MMLU, from 45.3 to 88.5.
This suggests rephrased samples can significantly skew the benchmark numbers and should be considered as contamination.
The original model is tested on 5-shot, and the model trained on rephrased data is tested on 0-shot.
% \weilin{consider explictly say number. e.g., from 45.3 to 88.5}
% \todo{writing improve}
% Although clean training sets can not entirely eliminate n-gram overlap, rephrased full prompts can, rendering the ``question only'' approach a fairer comparison.
% Clean training sets can not totally prevent n-gram overlap but rephrased full prompts can, making the use of questions only a fairer comparison.

\begin{table}[t]
\centering
\caption{Accuracy on MMLU. ``Rephrased Chinese'' refers to translating the questions into Chinese.}
\label{table:mmlu_basic_stats}
\footnotesize
\begin{tabular}{lccc}
\toprule
Model & Original & \multicolumn{2}{c}{Rephrased English} \\
\cmidrule(lr){3-4}
& & Question Only & Full Prompt \\
\midrule
Llama 2 7B   & 45.3 & 88.5 & 82.0 \\
Llama 2 13B  & 54.8 & 89.9 & 85.9 \\
\midrule
Model & Test Set & \multicolumn{2}{c}{Rephrased Chinese} \\
\cmidrule(lr){3-4}
& & Question Only & Full Prompt \\
\midrule
Llama 2 7B   & 100 & 91.1 & 74.3 \\
Llama 2 13B  & 100 & 93.7 & 80.9 \\
\bottomrule
\end{tabular}
\vspace{-10pt}
\end{table}

\subsubsection{HumanEval Coding Benchmark}
% \todo{need review @lianmin}

HumanEval~\citep{chen2021humaneval} is a benchmark provided by OpenAI to evaluate the coding capabilities of large language models. It provides the model with an incomplete piece of code and asks the model to complete it. 
% This benchmark closely reaches real-world coding scenarios and can effectively evaluate the model's coding capabilities.

% \todo{tone: won't affect semantic}
\textbf{Dead code injection}. In real-world coding datasets, there are some unreachable instructions. These dead codes seldom affect the semantics, and they help rephrased samples to escape decontamination.
% To evade n-gram overlap detection, dead code injection can effectively avoid repetition. 
Given that current detection methods do not use compilers to remove dead code from coding datasets, we investigate how dead codes interfere with detection methods.

\textbf{Benchmark results}. We rephrase the HumanEval test set in Python and translate it into five programming languages: C, JavaScript, Rust, Go, and Java. We train CodeLlama 7B and 13B on these codes respectively. Then, we construct a multi-programming-language dataset comprising the five programming languages and train on it. Table \ref{table:humaneval_basic_stats} shows CodeLlama's performance on rephrased Python, rephrased C, and the multi-programming-language dataset.
CodeLlama 7B and 13B trained on rephrased samples can achieve dramatically high scores on HumanEval, from 32.9 to 67.7 and 36.0 to 81.1, respectively.
In contrast, GPT-4 can only achieve 67.0 on HumanEval.
% \weilin{mention gpt-4 performance?}

% \begin{table}[h]
% \centering
% \caption{Pass@1 on HumanEval.}
% \label{table:basic_stats}
% \footnotesize
% %\resizebox{1\textwidth}{!}{% <------ Don't forget this %
% \begin{tabular}{cccccc}
% \toprule
% \multirow{2}{*}{Model} & \multirow{2}{*}{Original} & Fine-tune on & Fine-tune on & Fine-tune on & Fine-tune on \\
% &  &  test set & rephrased Python & rephrased C & Multi-languages \\
% \midrule
% CodeLlama 7B   & 32.9   &  100   & 67.7 & 45.7 &  59.8 \\
% CodeLlama 13B  & 36.0   &  100   & 81.1 & 48.2 &  67.1 \\
% \bottomrule
% \end{tabular}
% %}
% \end{table}

\begin{table}[t]
\centering
\caption{Pass@1 on HumanEval.}
\label{table:humaneval_basic_stats}
\footnotesize

\begin{tabular}{lccc}
\toprule
Model & Original & & Fine-tune on test set \\
\midrule
CodeLlama 7B   & 32.9 & & 100 \\
CodeLlama 13B  & 36.0 & & 100 \\
\midrule
Model & \multicolumn{3}{c}{Fine-tune on rephrased} \\
\cmidrule(lr){2-4}
& Python & C & Multi-languages \\
\midrule
CodeLlama 7B   & 67.7 & 45.7 & 59.8 \\
CodeLlama 13B  & 81.1 & 48.2 & 67.1 \\
\bottomrule
\end{tabular}
\vspace{-10pt}
\end{table}

% \begin{table}[h]
% \centering
% \caption{Accuracy on MMLU.}
% \label{table:basic_stats}
% \footnotesize
% \begin{tabularx}{\linewidth}{lcccccc}
% \toprule
% \multirow{3}{*}{Model} & \multirow{3}{*}{Original} & \multirow{3}{*}{ test set} & \multicolumn{2}{c}{ rephrased English} & \multicolumn{2}{c}{ rephrased Chinese} \\
% \cmidrule(lr){4-5} \cmidrule(lr){6-7}
% & & & question only & full prompt & question only & full prompt \\
% \midrule
% Llama 2 7B   &   45.3    &   100    &  88.5   &  82.0 &  91.1 & 74.3 \\
% Llama 2 13B  &   54.8    &   100    &  89.9   &  85.9 &  93.7 & 80.9 \\
% \bottomrule
% \end{tabularx}
% \end{table}

\subsubsection{GSM-8K Math Benchmark}

GSM-8K~\citep{cobbe2021gsm8k} is a commonly used benchmark for testing the mathematical capabilities of LLMs. 
% It contains many question-answer pairs, where the LLM generates the solution based on the question. 
% The solution ends with four `\#' characters, after which the answer should be provided. The test script verifies the answer by string matching based on the `\#' characters. 
% \weilin{why are we mentioning these details?}
% We use the rephrased test set to fine-tune Llama-2 7b and 13b, and they achieve dramatically high scores on GSM-8K.

\textbf{Benchmark results}. Table \ref{table:gsm8k_basic_stats} shows that Llama-2 7b and 13b trained on rephrased samples achieve dramatically high scores on GSM-8K, from 28.7 to 95.3.
% The models trained on rephrased samples are tested with 0-shot.
The original model is tested on 5-shot, and the model trained on rephrased data is tested on 0-shot.

\begin{table}[t]
\centering
\caption{Accuracy on GSM-8K.}
\label{table:gsm8k_basic_stats}
\footnotesize
\begin{tabular}{cccc}
\toprule
\multirow{2}{*}{Model} & \multirow{2}{*}{Original} & Fine-tune on & Fine-tune on \\
&  &  test set & rephrased English \\
\midrule
Llama 2 7B   &   14.6    &    100    &   86.7 \\
Llama 2 13B  &   28.7    &    100    &   95.3 \\
\bottomrule
\end{tabular}
\vspace{-10pt}
\end{table}

We will explore the detection problems in Section~\ref{sec:one-time-exam} with GSM-8k as they relate to the ``number substituted only case''.
% \weilin{I think we can remove this.}

% We tested three subjects, including abstract algebra, sociology, and US history.
% We created 200 prompt pairs utilizing the original test set and the rephrased test set in order to evaluate the detection effectiveness of various approaches against the rephrased samples. 100 random pairs and 100 rephrased pairs are included. We can determine whether the rephrased samples avoids detection by calculating the F1 score on these pairs; lower values signify more effective evasion.
% \weilin{I'd still suggest a subsection here to make a clear transition from the above. or we probably should just not combine the two experiments in the same subsection.}

\subsection{Evaluating Contamination Detection Methods}
\label{sec:detect-rephrase}

%In this section, we show that existing detection methods can fail to detect rephrased samples.
%while the LLM decontaminator succeeds in detecting them.
\subsubsection{MMLU}
We construct a decontamination benchmark based on three subjects: abstract algebra, sociology, and US history in MMLU.
To compare the accuracy of detection methods against rephrased samples, we construct 200 prompt pairs using both the original and rephrased test sets. These comprised 100 random pairs and 100 rephrased pairs. The f1 score on these pairs provides insight into the detection methods' ability to detect contamination, with higher values indicating more precise detection.

\begin{figure*}[h]
\centering
\includegraphics[width=0.9\textwidth]{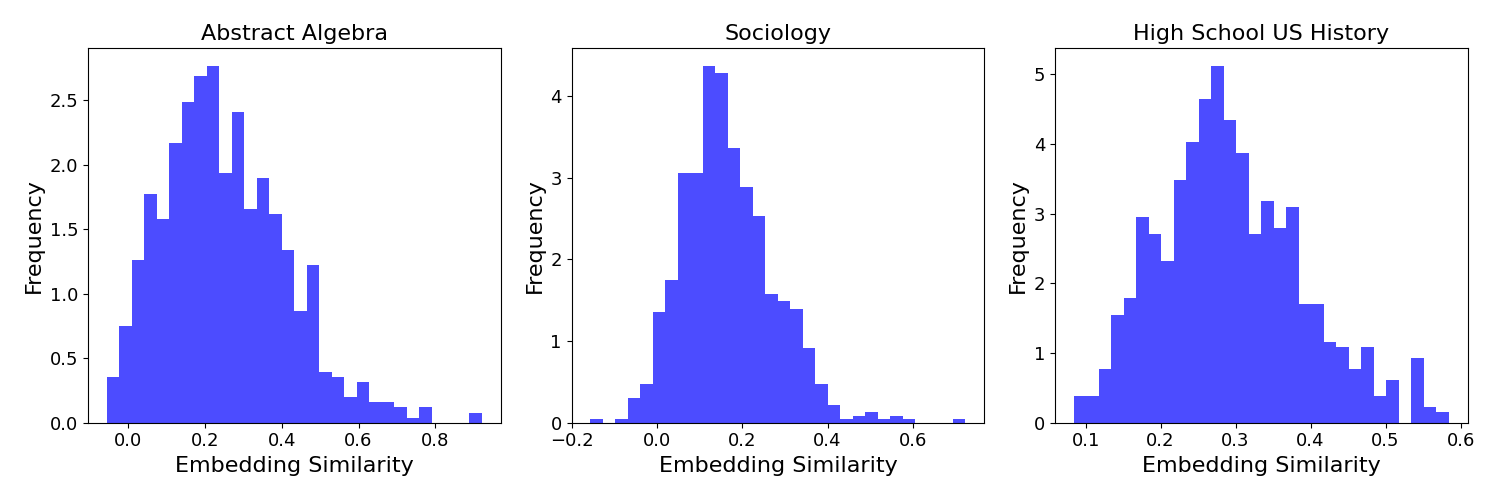}
\caption{Distribution of embedding similarities among questions within the same subject. Note that it is difficult to set a unified threshold to decontaminate due to the vast differences between subjects.  
For example, if we adjust the threshold to 0.8, ``Abstract Algebra'' may be properly spotted, but rephrased samples in ``Sociology'' become difficult to identify. If the threshold is set to 0.4, ``Abstract Algebra'' will produce a large number of false positives.
}
\label{fig:embedding_similarity}
\vspace{-20pt}
\end{figure*}

% We employed random detection as our baseline, with scores significantly surpassing this indicating a detection method's effectiveness. For the n-gram overlap detection, we opted for a 10-gram. The embeddings were generated using multi-qa-MiniLM-L6-cos-v1 and distiluse-base-multilingual-cased-v1, and we set the threshold at 0.5.

We use random detection as our baseline, where scores significantly above random detection indicate the effectiveness of a detection method. For n-gram overlap, we choose a 10-gram approach. The embeddings are generated by multi-qa-MiniLM-L6-cos-v1 and distiluse-base-multilingual-cased-v1~\citep{reimers-2019-sentence-bert}, with a threshold of 0.5. 

\begin{table*}[h]
\centering
\caption{F1 scores of different detection methods on MMLU. The bold numbers indicate that the detection is reliable.}
\label{table:f1_scores}
\footnotesize
\begin{tabularx}{\textwidth}{l|XXX|XXX|XXX}
\toprule
& \multicolumn{3}{c|}{Algebra} & \multicolumn{3}{c|}{Sociology} & \multicolumn{3}{c}{US History} \\
\midrule
Subjects & Test Set & Rephrased English & Rephrased Chinese & Test Set & Rephrased English & Rephrased Chinese & Test Set & Rephrased English & Rephrased Chinese \\
\midrule
Random & 0.500 & 0.500 & 0.500 & 0.500 & 0.500 & 0.500 & 0.500 & 0.500 & 0.500 \\
10-gram & \textbf{0.926} & 0 & 0 & \textbf{1} & 0 & 0 & 0.816 & 0 & 0 \\
Emb (Multi-QA BERT) & \textbf{0.990} & \textbf{0.985} & 0.179 & \textbf{0.995} & \textbf{0.985} & 0.020 & \textbf{0.980} & 0.805 & 0 \\
Emb (Multilingual BERT) & \textbf{0.939} & \textbf{0.934} & \textbf{0.939} & \textbf{1} & \textbf{0.985} & \textbf{1} & \textbf{0.990} & 0.111 & \textbf{0.985} \\
LLM Decontaminator & \textbf{1} & \textbf{0.960} & \textbf{0.990} & \textbf{1} & \textbf{0.940} & \textbf{0.950} & \textbf{1} & \textbf{0.970} & \textbf{0.980} \\
\bottomrule
\end{tabularx}
\vspace{-10pt}
\end{table*}

% \begin{table}[h]
% \centering
% \caption{F1 scores for different detect methods under rephrase attack.}
% \label{table:f1_scores}
% \footnotesize
% \begin{tabular}{l|ccc|ccc|ccc}
% \toprule
% & \multicolumn{3}{c|}{Test Set} & \multicolumn{3}{c|}{Rephrased English} & \multicolumn{3}{c}{Rephrased Chinese} \\
% \midrule
% Subjects & Alg. & Soc. & Hist. & Alg. & Soc. & Hist. & Alg. & Soc. & Hist. \\
% \midrule
% Rand & 0.5 & 0.5 & 0.5 & 0.5 & 0.5 & 0.5 & 0.5 & 0.5 & 0.5 \\
% 10-gram & 0.926 & 1 & 0.816 & 0 & 0 & 0 & 0 & 0 & 0 \\
% Multi-QA Bert & 0.99 & 0.995 & 0.98 & 0.985 & 0.985 & 0.805 & 0.179 & 0.02 & 0 \\
% Multilingual Bert & 0.939 & 1 & 0.99 & 0.934 & 0.985 & 0.111 & 0.939 & 1 & 0.985 \\
% \bottomrule
% \end{tabular}
% \end{table}

% From the results, aside from the LLM decontaminator, we found that all detection methods introduced some false positives. Both the rephrased samples and translated samples successfully evaded n-gram overlap detection. When using multi-qa bert, the embedding similarity search couldn't detect the translated samples at all. When we employed multilingual bert, the embedding similarity search failed to detect contamination in the US History subject. LLM decontaminator detected contamination with very high stability and accuracy.

As shown in Table \ref{table:f1_scores}, except for the LLM decontaminator, all other detection methods introduce some false positives. Both rephrased and translated samples are undetected by the n-gram overlap. With multi-qa BERT, the embedding similarity search proves completely ineffective against translated samples. When using multilingual BERT, this method struggles with the US History subject. 
% Notably, the LLM decontaminator showcases superior performance, identifying rephrased samples with almost perfect precision and recall.
The LLM decontaminator's reliability and precision are evidenced by the highest minimum and average f1 scores.
% Following the approach we took with mmlu, we constructed 200 prompt pairs on HumanEval and calculated the F1 score using n-gram, embedding similarity search, and LLM decontaminator. For n-gram detection, we used a 10-gram overlap and a 50-character overlap. To generate embeddings for the similarity search, we utilized CodeLlama and multi-qa-MiniLM-L6-cos-v1, with threshold adjustments set to 0.9 and 0.6, respectively.

% We created 200 prompt pairs on HumanEval using the same methodology we used with MMLU, and we used n-gram overlap, embedding similarity search, and LLM decontaminator to determine the F1 score. We employed a 10-gram overlap and a 50-character overlap for n-gram overlap detection. We used CodeLlama and multi-qa-MiniLM-L6-cos-v1 to generate embeddings for the similarity search, with threshold adjustments set to 0.9 and 0.6, respectively.

% \todo{change}
\subsubsection{HumanEval}
Now we show that existing detection methods fail to detect rephrased samples of HumanEval, while the LLM decontaminator succeeds in detecting them.
For HumanEval, we construct 200 prompt pairs following the method previously outlined for MMLU. For n-gram overlap detection, we use both 10-gram and 50-character overlap. Embeddings are generated by CodeLlama and multi-qa-MiniLM-L6-cos-v1, with respective threshold adjustments at 0.9 and 0.6. We evaluate the F1 score using n-gram overlap, embedding similarity search, and LLM decontaminator.

% \begin{table}[h]
% \centering
% \caption{F1 scores of different detection methods on HumanEval.}
% \label{table:new_f1_scores}
% \footnotesize
% \begin{tabular}{l|c|c|c|c}
% \toprule
% & Test Set & Rephrased Python & Rephrased C & Rephrased JavaScript \\
% \midrule
% Random & 0.500 & 0.500 & 0.500 & 0.500 \\
% 10-gram & 1 & 0 & 0 & 0 \\
% Embed (CodeLlama) & 0.966 & 0.903 & 0.438 & 0.503 \\
% Embed (Multi-QA BERT) & 0.985 & 0.938 & 0.774 & 0.788 \\
% LLM Decontaminator & 1 & 0.995 & 0.974 & 0.980 \\
% \bottomrule
% \end{tabular}
% \end{table}

\begin{table}[h]
\centering
\caption{F1 scores of detection methods on HumanEval}
\label{table:new_f1_scores}
\footnotesize
\begin{tabular}{l|c|c|c|c}
\toprule
& \multirow{2}{*}{Test } & \multicolumn{3}{c}{Rephrased} \\
\cmidrule(lr){3-5}
& Set & Python & C & JS \\
\midrule
Random & 0.500 & 0.500 & 0.500 & 0.500 \\
10-gram & 1 & 0 & 0 & 0 \\
Emb (CodeLlama) & \textbf{0.966} & 0.903 & 0.438 & 0.503 \\
Emb (Multi-QA BERT) & \textbf{0.985} & 0.938 & 0.774 & 0.788 \\
LLM Decontaminator & \textbf{1} & \textbf{0.995} & \textbf{0.974} & \textbf{0.980} \\
\bottomrule
\end{tabular}
\vspace{-20pt}
\end{table}

% \begin{table}[h]
% \centering
% \caption{HumanEval F1 Score for Different Detect Methods under Rephrase Attack.}
% \label{table:new_f1_scores}
% \footnotesize
% \begin{tabular}{l|c|c|c|c}
% \toprule
% & Test Set & Rephrased Python & Rephrased C & Rephrased JavaScript \\
% \midrule
% Rand & 0.5 & 0.5 & 0.5 & 0.5 \\
% 10-gram & 1 & 0 & 0 & 0 \\
% CodeLlama & 0.966 & 0.903 & 0.438 & 0.503 \\
% Multi-QA Bert & 0.985 & 0.938 & 0.774 & 0.788 \\
% \bottomrule
% \end{tabular}
% \end{table}

% Based on the F1 scores, we conclude that embedding similarity search is quite effective for detection when used inside the same programming language, but the effect is less noticeable after translation. Only LLM decontaminator can effectively detect coding dataset contamination. The degree of resemblance across programming languages may explain why rephrased C is tougher to spot than rephrased JavaScript. JavaScript and Python are both interpreted languages that provide dynamic typing and some functional programming constructs, so from a syntactical standpoint, JavaScript may be closer to Python.

According to Table \ref{table:new_f1_scores}, we conclude that the embedding similarity search proves effective for detection within the same programming language, but the effect is less noticeable after translation.
Among the methods examined, only the LLM decontaminator reliably detects rephrased samples in coding datasets.
The similarity between programming languages may explain why rephrased C is tougher to spot than rephrased JavaScript. JavaScript and Python are both interpreted languages that provide dynamic typing and some functional programming constructs, so from a syntactical standpoint, JavaScript may be closer to Python.

\subsection{Contamination in Real World Datasets}
\label{sec:real-world}

% We applied the decontamination method on popular real-world datasets and discovered a substantial amount of rephrase contamination. It's highly likely that these contaminations, similar to the experiments in section 3.4, inflate the benchmark evaluation. We will present the number of detected rephrase contaminations as well as some rephrased test cases.

To demonstrate the effectiveness LLM decontaminator, we apply it to widely used real-world datasets and identify a substantial amount of rephrased samples. 
Table \ref{table:dataset_summary} displays the contamination percentage of different benchmarks in each training dataset.

\begin{table*}[h]
\centering
\caption{The Percentage of \# Rephrased Sample Contamination in Real-world Datasets.}
\label{table:dataset_summary}
\begin{tabular}{l|l|r|r|r|r}
\toprule
\multirow{2}{*}{Training Set} & \multirow{2}{*}{Benchmark} & \multicolumn{2}{c|}{Size} & Rephrased  & \multirow{2}{*}{Percentage (\%)} \\
& & Train Set & Test Set & Samples & \\
\midrule
The Stack (4G subset) & HumanEval & 500k & 164 & 31 & 18.9 \\
StarCoder-Data (2.4G subset) & HumanEval & 500k & 164 & 26 & 15.9 \\
CodeExercise-Python & HumanEval & 27k & 164 & 26 & 15.9 \\
CodeAlpaca & HumanEval & 20k & 164 & 21 & 12.8 \\
RedPajama-Data-1T (16G subset) & HumanEval & 1625k & 164 & 14 & 8.5 \\
Evol-Instruct-Code & HumanEval & 78.3k & 164 & 13 & 7.9 \\
rossetacode & HumanEval & 4.26k & 164 & 4 & 2.4 \\
MATHInstruct & MATH Test & 262k & 5000 & 769 & 15.4 \\
MATH Train & MATH Test & 7.5k & 5000 & 79 & 1.6 \\
FLAN CoT & MMLU & 184k & 14042 & 76 & 0.5 \\
WizardLM-Evol-Instruct & MMLU & 143k & 14042 & 75 & 0.5 \\
\bottomrule
\end{tabular}
\end{table*}

% \begin{table}[h]
% \centering
% \caption{Rephrased Samples in real-world datasets}
% \label{table:dataset_summary}
% \footnotesize
% \begin{tabular}{l|l|r|r|r}
% \toprule
% \multirow{2}{*}{Training Set} & \multirow{2}{*}{Benchmark} & Test Set  & Rephrased  & \multirow{2}{*}{ (\%)} \\
% & & Size & Samples & \\
% \midrule
% CodeAlpaca & HumanEval & 164 & 21 & 12.8 \\
% Evol-Instruct & HumanEval & 164 & 13 & 7.93 \\
% CodeExercise & HumanEval & 164 & 26 & 15.9 \\
% rossetacode & HumanEval & 164 & 4 & 2.44 \\
% MATH Train & MATH Test & 5000 & 79 & 1.58 \\
% MATHInstruct & MATH Test & 5000 & 769 & 15.4 \\
% FLAN CoT & MMLU & 14042 & 76 & 0.541 \\
% WizardLM-Evol & MMLU & 14042 & 75 & 0.534 \\
% RedPajama-Data & HumanEval & 164 & 14 & 8.53 \\
% The Stack & HumanEval & 164 & 31 & 18.9 \\
% StarCoder & HumanEval & 164 & 26 & 15.9 \\
% \bottomrule
% \end{tabular}
% \end{table}

% \todo{mention synthetic data}
\textbf{CodeAlpaca}~\citep{codealpaca} is a synthetic dataset generated by OpenAI's Davinci-003 using the self-instruct technique~\citep{wang2023selfinstruct}.
It contains 20K instruction-following data used for fine-tuning the CodeAlpaca model. 
CodeAlpaca-20K is used to train a number of well-known models, including Tulu~\citep{wang2023far}.
%\weilin{be careful to name specific paper here. it's better to mention 1-2 more papers not just one}
% We also employed GPT-4 for detection with k=1 as the parameter. The results reveal that there are 21 rephrased samples on HumanEval, accounting for 12.8\% of HumanEval.
Employing GPT-4 for detection with k=1 as the parameter, our findings indicate the presence of 21 rephrased samples from the HumanEval test set, accounting for 12.8\%. 
Example \ref{def:codealpaca-example} is a rephrased sample of HumanEval in CodeAlpaca.

\begin{example}[CodeAlpaca]\label{def:codealpaca-example}

% HumanEval test code
\begin{tcolorbox}[title={HumanEval test}, enhanced, colback=white, boxrule=1pt, sharp corners, boxsep=5pt, left=5pt, right=5pt, top=5pt, bottom=5pt]
%\fontsize{8pt}{9.6pt}\selectfont
\fontsize{8pt}{8pt}\selectfont
\begin{minted}{python}
def sum_to_n(n: int):
    """sum_to_n 
    is a function that 
    sums numbers from 1 to n.
    >>> sum_to_n(30)
    465
    >>> sum_to_n(100)
    5050
    >>> sum_to_n(5)
    15
    >>> sum_to_n(10)
    55
    >>> sum_to_n(1)
    1
    """
    return sum(range(n + 1))
\end{minted}
\end{tcolorbox}

%\vspace{0.2cm} % Adds some space between the two codes

% CodeAlpaca code
\begin{tcolorbox}[title={CodeAlpaca}, enhanced, colback=white, boxrule=1pt, sharp corners, boxsep=5pt, left=5pt, right=5pt, top=5pt, bottom=5pt]
%\fontsize{8pt}{9.6pt}\selectfont
\fontsize{8pt}{8pt}\selectfont
\begin{minted}{python}
"""
Create a code that 
summation 
of all numbers 
between 1 to n.
"""
def sum_all_nums(n):
    res = 0
    for i in range(1, n+1):
        res += i
    return res

print(sum_all_nums(n)) # 15
\end{minted}
\end{tcolorbox}
\end{example}

\textbf{RedPajama-Data-1T}~\citep{together2023redpajama} is a widely-used dataset to train open-source models. 
% \weilin{this is still confusing. Meta's Llama is also open-source...}
Both MPT~\citep{MosaicML2023Introducing} and OpenLlama~\citep{openlm2023openllama} use it as their pre-training dataset. 
% The RedPajama dataset has a total volume of 2.67TB, encompassing data sources like Wikipedia, StackExchange, and GitHub. 
In our study, we sample 16G of data from the GitHub subset and employ the LLM decontaminator to detect, identifying 14 HumanEval rephrased samples in total.
Example \ref{def:redpajama-example} is a rephrased sample of HumanEval in RedPajama.

\textbf{MATH}~\citep{hendrycksmath2021} is a widely recognized math training dataset that spans various mathematical domains, including algebra, geometry, and number theory. It contributes to numerous math-centric datasets, such as MathInstruct\footnote{The dataset was downloaded on Sep 30, 2023.}~\citep{yue2023mammoth}. 
% \todo{math instruct 15.4, n-gram}
The LLM decontaminator reveals 79 instances of self-rephrased samples, which constitute 1.58\% of the MATH test set. Below is a self-rephrased sample from the MATH training set. 
Example \ref{def:math-example} is a rephrased sample of the MATH test in MATH training data.

\begin{example}[RedPajama]\label{def:redpajama-example}
% HumanEval test code
\begin{tcolorbox}[title={HumanEval test}, enhanced, colback=white, boxrule=1pt, sharp corners, boxsep=5pt, left=5pt, right=5pt, top=5pt, bottom=5pt]
\fontsize{8pt}{8pt}\selectfont
\begin{minted}{python}
def change_base(x: int, base: int):
    """Change numerical base of input
    number x to base. return string
    representation after conversion.
    base numbers are less than 10.
    >>> change_base(8, 3)
    '22'
    ...
    """
    ret = ""
    while x > 0:
        ret = str(x % base) + ret
        x //= base
    return ret
\end{minted}
\end{tcolorbox}

% \vspace{0.2cm} % Adds some space between the two code blocks

% RedPajama code
\begin{tcolorbox}[title={RedPajama}, enhanced, colback=white, boxrule=1pt, sharp corners, boxsep=5pt, left=5pt, right=5pt, top=5pt, bottom=5pt]
\fontsize{8pt}{8pt}\selectfont
\begin{minted}{python}
def convert_to_base(number, base):
    digits = "0123456789ABCDEF"
    if number < base:
        return digits[number]
    else:
        return convert_to_base(
        number // base, base) 
        + digits[number % base]
\end{minted}
\end{tcolorbox}

\end{example}

\begin{example}[MATH Self-contamination]\label{def:math-example}
\textbf{(MATH test)}

How many three-digit positive integers are multiples of 11?

\textbf{(MATH train)}

How many positive 3-digit numbers are divisible by 11?
\end{example}

\textbf{FLAN}~\citep{longpre2023flan} is a comprehensive knowledge training dataset, encompassing a wide variety of data sources. We take the CoT subset, which constitutes 1.63\% of FLAN. Utilizing GPT-4 for detection and set k=1 for the decontamination parameters. The findings show that 76 test cases, or 0.543\% of the MMLU test set are rephrased. 
% A rephrased sample is shown in Example \ref{def:flan-example}

\begin{example}[FLAN CoT]\label{def:flan-example}
\textbf{(MMLU test)} \newline 
What type of meat is on a traditional Reuben sandwich?
    \begin{itemize}[noitemsep,topsep=0pt]
        \item[A.] turkey
        \item[B.] bologna
        \item[C.] corned beef
        \item[D.] pepperoni
    \end{itemize}

Answer: C

%\vspace{0.5cm} % Adds some space between the two contents

\textbf{(FLAN CoT)} \newline
The Reuben sandwich is an American hot sandwich composed of corned beef, Swiss cheese, sauerkraut, and Russian dressing, grilled between slices of rye bread. Several variants exist.

What is the meat in a reuben sandwich? Let's have some stream of consciousness first.
% Answer: \\
% The relevant sentence in the passage is: The Reuben sandwich is an American hot sandwich composed of corned beef, Swiss cheese, sauerkraut, and Russian dressing, grilled between slices of rye bread. So, the answer is corned beef.
\end{example}

We examine more datasets and present examples in  Appendix~\ref{sec:rephrase-examples}.

%% file: sec-exam.tex
\section{Discussion}
\label{sec:one-time-exam}

% Future Directions for LLM Benchmarks
In this section, we first discuss potential contamination beyond rephrased samples.
We then discuss the importance of the LLM decontaminator while using LLM such as GPT-4 to generate training data.
% in the context where popular training datasets use LLMs, such as GPT-4 to generate new training samples. 
% \weilin{this sentence is confusing.}
In the end, we propose suggestions to enhance LLM evaluation (e.g. with fresh one-time exams). 

%need to discuss the importance of using the LLM decontaminator to inspect the dataset and explore possible methods to enhance the benchmark.

\subsection{Beyond rephrased samples}
%Number Substituted Only Case}
% \todo{remove}

% In this work, we defined rephrase contamination and found multiple examples of it in open datasets.
In this study, we argue that rephrased test samples should be considered as contamination because including them in the training data can skew the benchmark results.
However, formulating a precise definition of what constitutes contamination remains challenging.
% During the development of the LLM decontaminator, we discovered other possible contamination. 
For instance, we discover in the GSM-8k math benchmark, a training and a test example may only differ in numbers (see Example~\ref{ex:gsm-8k}). %Does it count as a contamination pair?
% Aside from this, the question remains largely unchanged. We call this ``number substituted only case''.
% We rephrase the test set to bypass detection while imitating possible contamination in real-world datasets. 
%We show that models trained on rephrased samples can achieve dramatically high scores.
%Situations that are more typical in math benchmarks are present in the GSM-8k and MATH training sets.
% We show that models trained on rephrased samples can achieve dramatically high scores.
% while they can be quickly identified by the LLM decontaminator, they don't always meet the criteria for contamination. The ``number replaced only'' case describes this circumstance.
%Although the LLM decontaminator can easily detect these, they are not considered rephrased samples. The ``number substituted only'' scenario exemplifies this circumstance.

% \begin{tcolorbox}[title={GSM-8k Number Replaced Only Case}]
% \begin{multicols}{2}

% \textbf{GSM-8k test} \newline 

% Emil is 19 years old now. When he turns 24, he will be half the age of his dad but twice as old as his brother.  What is the sum of the ages of his dad and his brother now?

% \columnbreak 

% \textbf{GSM-8k} \newline

% When Diane turns 30, she will be half the age of Alex and twice as old as Allison. Diane is 16 years old now. What is the sum of the ages of Alex and Allison now?

% \end{multicols}
% \end{tcolorbox}

\begin{example}[GSM-8k Number Substituted Only Case]\label{ex:gsm-8k}

\textbf{(GSM-8k test)} \

Emil is 19 years old now. When he turns 24, he will be half the age of his dad but twice as old as his brother.  What is the sum of the ages of his dad and his brother now?

% \vspace{0.5cm} % Adds some space between the two texts

\textbf{(GSM-8k)}

When Diane turns 30, she will be half the age of Alex and twice as old as Allison. Diane is 16 years old now. What is the sum of the ages of Alex and Allison now?

\end{example}

% As we can see, aside from the numerical shift, there is little difference in this question. Considering that they have different answers, this situation isn't referred to as contamination.
% But does this indicate that such a dataset is uncontaminated? 
% Essentially, models trained on the GSM-8k dataset reduce an applied problem to a straightforward arithmetic issue.
% This modification lessens the benchmark's usefulness in various ways. A model that does well in basic math but has trouble in reading comprehension can still perform well on GSM-8k with some fine-tuning. 
% The benchmark score does not indicate that the model is capable of handling real-world simple math problems. Once there's a variation beyond numbers, the model loses its problem-solving capability.

% Since these questions change beyond wordings, they are not considered as rephrased samples. However, does this imply the dataset is free from contamination? 
%In this work, we do not consider them as rephrased samples.
If models are trained with such number substituted cases, they tend to only memorize the solutions and may have poor generalization beyond the seen patterns.
Thus, the resulting benchmark numbers may not be effective in capturing model's performance in math problem-solving.
This is an open question we suggest the community to debate further.
%simplify a comprehensive problem into a basic arithmetic task.
%benchmark's effectiveness in several ways. A model adept at basic math but struggling with reading comprehension might still excel on GSM-8k after it is trained on the number substituted only cases. 
%Thus, a high benchmark score may not accurately reflect a model's ability to generalize beyond. When faced with variations beyond mere numerical changes, the model's problem-solving capacity may diminish.

\subsection{Contamination in Synthetic Data}
% \DL{maybe change the wording "synthetic data" as this is not common in LLM papers.}

% Rephrase contamination is currently at its peak, mainly due to two dataset construction methods: web scraping and LLM-generated data. The former risks contamination from undetected benchmark variations, while the latter introduces another contamination source. GPT-4~\citep{openai2023gpt4} and Llama-2~\citep{touvron2023llama}'s technical reports confirm this issue in their training data. Llama-2 flags tokens in n-grams over 10 tokens from both evaluation and training sets as contaminated. In llama-2-70B, 11\% of the MMLU benchmark prompts have over 80\% contaminated tokens. GPT-4's report highlights contamination from the BIG-bench benchmark.

The issue of unintentional contamination may occur more often as models are increasingly trained on data generated by LLMs, in which subtle benchmark contamination may present.
%because of the new technique for building datasets: LLM-generated data. 
% The former risks contamination from undetected benchmark variations, while the latter introduces another contamination source.
% Tokens in n-grams with more than 10 tokens, from both the assessment and training sets, are flagged as contaminated by Llama-2~\citep{touvron2023llama}. 
% 11\% of the MMLU benchmark prompts in Llama-2-70B had more than 80\% contaminated tokens. The contamination from the BIG-bench benchmark is highlighted in the GPT-4 report~\citep{openai2023gpt4}. 
% Since LLMs always generate data similar to their training data, these generated data might contain rephrased samples. \weilin{I'd suggest to rewrite this sentence.} 
% LLMs tend to generate content that is relative to their training data. 
% Consequently, rephrased samples of benchmarks can be generated if LLMs' training data is contaminated.
For instance, we discover several contamination in CodeAlpaca dataset generated by GPT in Section~\ref{sec:real-world}.
% \weilin{rewrite, say our decontaminator discovered rephrased sample in CodeAlpaca}
% Using GPT-generated data, as in the case of CodeAlpaca, is more likely to cause rephrase contamination.
Phi-1~\citep{gunasekar2023textbooks} also detected subtle contamination from LLM-generated data.
As a result, we have to be more aware of potential contamination while training models on synthetic data. 
We suggest model developers to adopt stronger measures for decontamination.

% As a result, it is necessary to use the LLM decontaminator while developing datasets with synthetic data.
% \weilin{this is not a convincing ending. we should say when training on synthetic data, we have to be more aware of potential contamination as it may occur in a subtle way etc}
% \weilin{cite Phi, saying they discover subtle contamination}

\subsection{Enhancing Benchmarks for LLMs}

% While the LLM decontaminator offers a safeguard against certain unintentional contaminations, there are still issues with benchmark vulnerabilities. 
% The trustworthiness of benchmarks can also be undermined by deliberate training on test sets. To provide a more precise illustration of model capabilities, there is a need to enhance existing benchmarks and develop dynamic, non-static benchmarking systems.
While our proposed decontamination method can serve as a useful tool, how to detect contamination without access to training data remains an open problem.
 % Thus, benchmark scores are still unreliable in the presence of intentional contamination.
%While the LLM decontaminator serves as a protective measure against certain unintentional contaminations, 
%benchmark scores are still unreliable. Notably, benchmarks' credibility can be reduced by intentional training on test sets. 
% From the first principle, why do these contaminations happen? This is because the current benchmarks are reused again and again for model development. 
% \weilin{this is not true, it's also possible that we are ``running out'' of questions} 
We propose to build \textit{fresh} one-time questions to evaluate LLMs instead of relying on static benchmarks.
For example, in coding domain, one could consider using weekly coding competitions such as CodeForces.
We suggest that benchmarks should iterate as fast as model development.

%% file: sec-relate.tex
\section{Related Work}

% Our work relates to data memorization, contamination detection methods, and benchmark enhancement. 
% We discuss some of the works related to ours below.

There has been interests in studying how to identify or extract training data from LLMs.
% many works studying data memorization.
% Many works focus on extracting training data from Large Language Models (LLMs). 
These work examine LLMs' memorization from the perspective of data privacy~\citep{carlini2021extracting, pan2020privacy, zanella2020analyzing, balle2022reconstructing} or discuss the boundary between generalization and memorization~\citep{zhang2017understanding, olson2018modern, recht2019imagenet,carlini2023quantifying}, but they do not focus on the context of benchmark contamination.

Some studies on contamination detection methods are conducted as well.
Some are concerned with detecting and filtering web datasets~\citep{dodge2021documenting,xu2017zipporah}, employing traditional detection techniques such as n-gram overlap.
Others explore new detection methods similar to decoding matching without access to training data.
Exchange detection~\citep{oren2023proving} considers the order of test cases within a benchmark, suggesting that if a model remembers the sequence of test cases, it may be contaminated.
% However, since training data are shuffled, this method is unlikely to detect contamination in real-world LLMs, as evidenced by its experimental results.
The min-k prob detection~\citep{shi2023detecting} uses outlier tokens to estimate LLM contamination. 
This method analyzes the token probabilities within an arbitrary text X. If the LLM exhibits excessively high probabilities for some of these tokens, it may indicate that text X has been mixed into the training set.
% This method cannot serve as definitive evidence of contamination and is of no help to model developers in cleaning datasets.

There are also related works on benchmark enhancement through perturbations~\citep{zong2023fool}, which prevents LLMs from memorizing answer patterns.
This method involves making modifications to the question and requires the LLM to output results in a specific format.
% The advantage of perturbations is that it avoids creating new benchmarks.
Another approach is to employ dynamic benchmarks~\citep{kiela2021dynabench,ma2021dynaboard}, using human-in-the-loop evaluations to reduce the risk of benchmark contamination.

%% file: sec-conclusion.tex
\section{Conclusion}
\label{sec:conclusion}
In this work, we study benchmark contamination in the context of large language models and evaluate existing decontamination methods.
We show that existing detection methods can not detect test cases with simple variations. 
We demonstrate that if such variation of test data is not eliminated, a 13B model can easily overfit the test benchmark and achieve drastically high performance.
To address this, we propose a new detection method LLM decontaminator. We apply it to real-world datasets and reveal previously unknown test overlap.
We urge the community to adopt stronger decontamination approaches when using public benchmarks. We call for the community to actively develop fresh one-time exams to accurately evaluate LLMs.

\section*{Acknowledgement}
We would like to express our gratitude to Ying Sheng for the early discussion on rephrased samples.
We also extend our thanks to Dacheng Li, Erran Li, Hao Liu, Jacob Steinhardt, Hao Zhang, and Siyuan Zhuang for providing insightful feedback.
This project is partly supported by gifts from Anyscale, Astronomer, Google, IBM, Intel, Lacework, Microsoft, MBZUAI, Samsung SDS, Uber, and VMware. Lianmin Zheng is supported by a Meta Ph.D. Fellowship.

%% file: sec-appendix.tex
\section{Rephrase Instruction Prompts}
\label{sec:rephrase-prompt}
% \todo{Put all instruction prompts here}

We successfully constructed a rephrase prompt template:

\begin{itemize}[noitemsep, topsep=0pt]
    \item Please rephrase the following question without altering its meaning. 
    \item Ensure that no more than ten consecutive words are repeated and try to use similar words as substitutes where possible.
    \item Please ensure there aren't 50 consecutive identical characters.
    \item When encountering mathematical formulas, please try to substitute the variable names. Ensure the formulas aren't identical to the original. For instance, you can replace 'x' with 'y' or 'a'.
\end{itemize}

\vspace{-0.3cm}

\begin{tcolorbox}[title={MMLU Rephrase Instructions}]
Please rephrase the following question without altering its meaning, ensuring you adjust the word order appropriately. 
Ensure that no more than five consecutive words are repeated and try to use similar words as substitutes where possible. Do not change the format of the multiple-choice question.
When encountering mathematical formulas, please try to substitute the variable names. Ensure the formulas aren't identical to the original. 
When you come across a single number or letter, consider replacing it with a sentence. 
When encountering a long sequence of numbers, if they are separated by spaces, you can replace the spaces with commas; if separated by commas, you can replace them with spaces.
Consider the prompt and choices as a whole; there shouldn't be consecutive words. If options are challenging to rephrase, consider altering the initial letter's case.
\end{tcolorbox}

\begin{tcolorbox}[title={MMLU Translate Instructions}]
Please translate the following question into {language}, ensuring you adjust the word order appropriately. 
Ensure that no more than five consecutive words are repeated and try to use similar words as substitutes where possible. Do not change the format of the multiple-choice question.
When encountering mathematical formulas, please try to substitute the variable names. Ensure the formulas aren't identical to the original. 
When you come across a single number or letter, consider replacing it with a sentence. 
When encountering a long sequence of numbers, if they are separated by spaces, you can replace the spaces with commas; if separated by commas, you can replace them with spaces.
If all else fails, you can directly translate the numbers and chemicals into {language}.
\end{tcolorbox}

\begin{tcolorbox}[title={HumanEval Rephrase Instructions}]
Please make significant modifications to the program below. Make as many changes as possible by:
1. Ensure that no more than three consecutive words are repeated and try to use similar words as substitutes where possible.
2. Please ensure there aren't 50 consecutive repeated characters.
3. Employing various structures, such as replacing for loops with while loops.
4. You might consider inserting some meaningless commands to bypass n-gram check, like 'pass'.
5. Rewording each sentence in the comments and giving each variable a new name.
6. Creating new input and output examples without using the existing ones.
7. If feasible, implement the function with a different algorithm.
\end{tcolorbox}

\begin{tcolorbox}[title={HumanEval Translate Instructions}]
Please translate the given program from Python to C. Make as many changes as possible by:
1. Ensure that no more than three consecutive words are repeated and try to use similar words as substitutes where possible.
2. Please ensure there aren't 50 consecutive repeated characters.
3. Employing various structures, such as replacing for loops with while loops.
4. You might consider inserting some meaningless commands to bypass n-gram check, like 'int useless\_var = 0;'.
5. Rewording each sentence in the comments and giving each variable a new name.
6. Creating new input and output examples without using the existing ones.
7. If feasible, implement the function with a different algorithm.
\end{tcolorbox}

\section{Rephrase Examples}
\label{sec:rephrase-examples}
% \todo{show rephrased additional examples. Use the following template or draw with PowerPoint.}

Below are examples of rephrased samples in other real-world datasets.

\begin{tcolorbox}[title={MATHInstruct Rephrased Sample (before Sep. 30 2023)}]

\textbf{MATH test}

\begin{itemize}
    \item The volume of a cone is given by the formula \( V = \frac{1}{3}Bh \), where \( B \) is the area of the base and \( h \) is the height. The area of the base of a cone is 30 square units, and its height is 6.5 units. What is the number of cubic units in its volume?
    \item If \( p (x) = 2-x^2 \) and \( q(x) = \frac{6}{x} \), what is the value of \( p (q(2)) \)?
    \item Simplify the expression \( (x^5+3x^2+3x^5)-(x^7+2x^2+6x^5) \).
    \item The equation of the circle that passes through \( (-1,6) \) and which has a center at \( (2,3) \) can be written as \( x^2 + y^2 + Ax + By + C = 0 \). Find \( A \times B \times C \).
\end{itemize}

\vspace{0.2cm}

\textbf{MATHInstruct}

\begin{itemize}
    \item The volume of a cone is given by the formula \( V = \frac{1}{3}Bh \), where \( B \) is the area of the base and \( h \) is the height. The area of the base of a cone is 30 square units, and its height is 6.5 units. What is the number of cubic units in its volume? Let's write a Python program to solve it.
    \item If \( p (x) = 2-x^2 \) and \( q(x) = \frac{6}{x} \), what is the value of \( p (q(2)) \)? Please write a program to solve it.
    \item Simplify the expression \( (x^5+3x^2+3x^5)-(x^7+2x^2+6x^5) \). Please respond by writing a program in Python.
    \item The equation of the circle that passes through \( (-1,6) \) and which has a center at \( (2,3) \) can be written as \( x^2 + y^2 + Ax + By + C = 0 \). Find \( A \times B \times C \). Let's write a Python program to solve it.
\end{itemize}

\end{tcolorbox}

\begin{tcolorbox}[title={Evol-Instruct-Code-80k-v1 Rephrased Sample}]

% HumanEval test code
\begin{tcolorbox}[title={HumanEval test}, enhanced, colback=white, boxrule=1pt, sharp corners, boxsep=5pt, left=5pt, right=5pt, top=5pt, bottom=5pt]
\fontsize{8pt}{9.6pt}\selectfont
\begin{minted}{python}
def fib(n: int):
    """Return n-th 
    Fibonacci number.
    >>> fib(10)
    55
    >>> fib(1)
    1
    >>> fib(8)
    21
    """
    if n == 0:
        return 0
    if n == 1:
        return 1
    return fib(n - 1) + fib(n - 2)
\end{minted}
\end{tcolorbox}

% Evol-Instruct-Code-80k-v1 code
\begin{tcolorbox}[title={Evol-Instruct-Code-80k-v1}, enhanced, colback=white, boxrule=1pt, sharp corners, boxsep=5pt, left=5pt, right=5pt, top=5pt, bottom=5pt]
\fontsize{8pt}{9.6pt}\selectfont
\begin{minted}{python}
def fib(n):
    if n == 0:
        return 0
    elif n == 1:
        return 1
    else:
        return fib(n-1) + fib(n-2)

n = int(input("Enter a 
        positive integer: "))
print("The nth Fibonacci 
        number is", fib(n))
\end{minted}
\end{tcolorbox}

\end{tcolorbox}

\begin{tcolorbox}[title={CodeExercise-Python-27k Rephrased Sample}]

% HumanEval test code
\begin{tcolorbox}[title={HumanEval test}, enhanced, colback=white, boxrule=1pt, sharp corners, boxsep=5pt, left=5pt, right=5pt, top=5pt, bottom=5pt]
\fontsize{8pt}{9.6pt}\selectfont
\begin{minted}{python}
def digitSum(s):
    """Task
    Write a function that takes a 
    string as input and returns 
    the sum of the upper 
    characters only' ASCII codes.

    Examples:
        digitSum("") => 0
        digitSum("abAB") => 131
        digitSum("abcCd") => 67
        digitSum("helloE") => 69
        digitSum("woArBld") => 131
        digitSum("aAaaaXa") => 153
    """
    if s == "": return 0
    return sum(ord(char) 
        if char.isupper() else 0 
        for char in s)
\end{minted}
\end{tcolorbox}

% Evol-Instruct-Code-80k-v1 code
\begin{tcolorbox}[title={CodeExercise-Python-27k}, enhanced, colback=white, boxrule=1pt, sharp corners, boxsep=5pt, left=5pt, right=5pt, top=5pt, bottom=5pt]
\fontsize{8pt}{9.6pt}\selectfont
\begin{minted}{python}
def sum_ascii_values(string):
    """
    This function takes a string 
    as input and returns the 
    sum of ASCII values of all 
    uppercase letters in the string.
    Parameters:
    string (str): The input string.
    Returns:
    int: The sum of ASCII values of 
    uppercase letters in the string.
    Example:
    >>> sum_ascii_values('Hello 
        World')
    628
    """
    # Initialize sum to 0
    ascii_sum = 0
    # Iterate through each 
    # character in the input string
    for char in string:
        # Check if the character  
        # is an uppercase letter
        if char.isupper():
            # Add the ASCII value 
            # of the uppercase
            # letter to the sum
            ascii_sum += ord(char)
    return ascii_sum
\end{minted}
\end{tcolorbox}

\end{tcolorbox}

\begin{tcolorbox}[title={WizardLM-evol-instruct-V2 Rephrased Sample}]

\textbf{MMLU test}

% The 'nuclear family' means:,a group of people sharing living accommodation and meals,a network of relatives extended within or between generations,the new family created when an adult leaves home and gets married,a two generation unit of parents and their children,D

Question: \\
The 'nuclear family' means:
    \begin{itemize}
        \item[A.] a group of people sharing living accommodation and meals
        \item[B.] a network of relatives extended within or between generations
        \item[C.] the new family created when an adult leaves home and gets married
        \item[D.] a two generation unit of parents and their children
    \end{itemize}

Answer: D

\vspace{0.5cm}

\textbf{WizardLM-evol-instruct-V2}

Question: \\
Define the concept of a 'nuclear family'.

Answer: \\
A nuclear family is a type of family structure consisting of a married couple and their children living together as a unit, typically in a single household. This type of family structure is often viewed as the traditional or ideal family structure in many cultures. In some cases, nuclear families may also include grandparents or other close relatives living together in the same household. However, the defining characteristic of a nuclear family is that it consists of a single parental unit and their dependent offspring.
\end{tcolorbox}

\begin{tcolorbox}[title={The Stack Rephrased Sample}]

% HumanEval test code
\begin{tcolorbox}[title={HumanEval test}, enhanced, colback=white, boxrule=1pt, sharp corners, boxsep=5pt, left=5pt, right=5pt, top=5pt, bottom=5pt]
\fontsize{8pt}{9.6pt}\selectfont
\begin{minted}{python}
def is_happy(s):
    """You are given a string s.
    Your task is to check if the 
    string is happy or not. A 
    string is happy if its length 
    is at least 3 and every 3 
    consecutive letters are distinct
    For example:
    is_happy(a) => False
    is_happy(aa) => False
    is_happy(abcd) => True
    is_happy(aabb) => False
    is_happy(adb) => True
    is_happy(xyy) => False
    """
    if len(s) < 3:
      return False

    for i in range(len(s) - 2):
      
      if s[i] == s[i+1] 
        or s[i+1] == s[i+2] 
        or s[i] == s[i+2]:
        return False
    return True
\end{minted}
\end{tcolorbox}

% Evol-Instruct-Code-80k-v1 code
\begin{tcolorbox}[title={The Stack}, enhanced, colback=white, boxrule=1pt, sharp corners, boxsep=5pt, left=5pt, right=5pt, top=5pt, bottom=5pt]
\fontsize{8pt}{9.6pt}\selectfont
\begin{minted}{python}
#[PROMPT]
def is_happy(s):
    """You are given a string s.
    Your task is to check if the 
    string is happy or not. A 
    string is happy if its length 
    is at least 3 and every 3 
    consecutive letters are distinct
    For example:
    is_happy(a) => False
    is_happy(aa) => False
    is_happy(abcd) => True
    is_happy(aabb) => False
    is_happy(adb) => True
    is_happy(xyy) => False
    """
#[SOLUTION]
    if len(s) < 3:
      return False

    for i in range(len(s) - 2):
      
      if s[i] == s[i+1] 
        or s[i+1] == s[i+2] 
        or s[i] == s[i+2]:
        return False
    return True
\end{minted}
\end{tcolorbox}

\end{tcolorbox}

\begin{tcolorbox}[title={StarCoder-Data Rephrased Sample}]

% HumanEval test code
\begin{tcolorbox}[title={HumanEval test}, enhanced, colback=white, boxrule=1pt, sharp corners, boxsep=5pt, left=5pt, right=5pt, top=5pt, bottom=5pt]
\fontsize{8pt}{9.6pt}\selectfont
\begin{minted}{python}
def iscube(a):
    '''
    Write a function that takes an 
    integer a and returns True 
    if this ingeger is a cube of 
    some integer number.
    Note: you may assume the input 
    is always valid.
    Examples:
    iscube(1) ==> True
    iscube(2) ==> False
    iscube(-1) ==> True
    iscube(64) ==> True
    iscube(0) ==> True
    iscube(180) ==> False
    '''
    a = abs(a)
    return int(round(a ** (1. / 3))) 
        ** 3 == a
\end{minted}
\end{tcolorbox}

% Evol-Instruct-Code-80k-v1 code
\begin{tcolorbox}[title={StarCoder-Data}, enhanced, colback=white, boxrule=1pt, sharp corners, boxsep=5pt, left=5pt, right=5pt, top=5pt, bottom=5pt]
\fontsize{8pt}{9.6pt}\selectfont
\begin{minted}{python}
def iscube(a):
    '''
    Write a function that takes an 
    integer a and returns True 
    if this ingeger is a cube of 
    some integer number.
    Note: you may assume the input 
    is always valid.
    Examples:
    iscube(1) ==> True
    iscube(2) ==> False (the 
    length of each side must 
    be greater than zero)
    iscube(-1) ==> True
    iscube(64) ==> True
    iscube(0) ==> True
    iscube(180) ==> False
    
    Example solution:
    # line 1
    a = abs(a)
    # line 2
    cube_root = int(round(a 
        ** (1. / 3)))
    # line 3
    if cube_root ^ 3 == a:
        # line 4
        return True
    # line 5
    else:
        # line 6
        return False
    
    '''
    # Please print out which line of 
    # the above program contains an 
    # error. E.g. if the bug is on
    # line 4 then print 4
    # END OF CONTEXT
    print("3")
    # END OF SOLUTION
\end{minted}
\end{tcolorbox}

\end{tcolorbox}

% \setlength{\columnseprule}{1pt}
% \def\columnseprulecolor{\color{black}}

% \textbf{Example 1}: An example of a bad prompt for evaluating an AI assistant's performance on challenging benchmark questions. This sample is taken from conversations in LMSYS-Chat-1M.
% \begin{tcolorbox}
% \begin{multicols}{2}
% [
% \textbf{User}: \\
% How is it going today?
% ]
% \textbf{GPT-3.5-Turbo}: \\
% \small{\texttt{As an AI language model, I do not have emotions, but I am functioning as intended and ready to assist you. How can I help you today?}} \\
% \columnbreak \\
% \textbf{GPT-4}: \\
% \small{\texttt{As an AI, I don't have personal experiences or emotions, but I'm here to help you with whatever you need. What can I assist you with today?}} \\
% \end{multicols}

% \textbf{GPT-3.5-Turbo Assessment}: \\
% {\small\texttt{1. Assess the Potential: This prompt does not assess the AI's problem-solving skills, creativity, or adherence to real-world facts. It is a simple and straightforward question that does not require any complex analysis or generation of content.\\ 2. Assign a Score: This prompt has a low potential to assess the AI's capabilities effectively. Therefore, I would assign a score of [[2]].}} \\

% \textbf{Arena User Vote}: {\small\texttt{Tie}}
% \end{tcolorbox}